\documentclass[sigconf, authorversion]{acmart}

\AtBeginDocument{%
  \providecommand\BibTeX{{%
    \normalfont B\kern-0.5em{\scshape i\kern-0.25em b}\kern-0.8em\TeX}}}

\acmConference[Preprint]{}{July 2022}{}

\usepackage{array}
\usepackage{amsmath}
\usepackage{amsfonts}
\usepackage[switch]{lineno}
\usepackage{subfigure}
\begin{document}

\title{Personalized Detection of Cognitive Biases in Actions of Users from Their Logs: Anchoring and Recency Biases}
%\title{Scoring the Dynamics of Users' Group Decision without Observing the Interactions among Users}
%\title{Predicting Collaborative Decision of a Group of Users without Observing  the Interactions among Users}
%Inferring Group Decision without Data on Members' Interactions

% SAMPLE
\author{Atanu R Sinha}
\authornote{Both authors contributed equally to this research.}
\email{atr@adobe.com}
\affiliation{%
  \institution{Adobe Research}
%   \city{Bangalore}
  \country{India}
}
\author{Navita Goyal}
\authornotemark[1]
\authornote{The work was done while the authors were at Adobe Research.}
\email{navita@umd.edu   }
\affiliation{%
  \institution{University of Maryland}
%   \city{College Park}
  \country{United States}
}
\author{Sunny Dhamnani}
\authornotemark[2]
\email{dhamnanisunny.1402@gmail.com}
\affiliation{%
  \institution{Meta Platforms}
  \country{United States}
}
\author{Tanay Asija}
\authornotemark[2]
\email{tanay.asija@gmail.com}
\affiliation{%
  \institution{Carnegie Mellon University}
  \country{United States}
}
\author{Raja K Dubey}
\authornotemark[2]
\email{rkdubey1677@gmail.com}
\affiliation{%
  \institution{Gan Studio}
  \country{India}
}
\author{M V Kaarthik Raja}
\authornotemark[2]
\email{kaarthikrajamv@gmail.com}
\affiliation{%
  \institution{Microsoft}
  \country{India}
}
\author{Georgios Theocharous}
\email{theochar@adobe.com}
\affiliation{%
  \institution{Adobe Research}
  \country{United States}
}

\renewcommand{\shortauthors}{Sinha and Goyal, et al.}

\begin{abstract}
Cognitive biases are mental shortcuts humans use in dealing with information and the environment, and which result in biased actions and behaviors (or, actions), unbeknownst to themselves. Biases take many forms, with cognitive biases occupying a central role that inflicts fairness, accountability, transparency, ethics, law, medicine, and discrimination. Detection of biases is considered a necessary step toward their mitigation. Herein, we focus on two cognitive biases - anchoring and recency. The recognition of cognitive bias in computer science is largely in the domain of information retrieval, and bias is identified at an aggregate level with the help of annotated data. Proposing a different direction for bias detection, we offer a principled approach along with Machine Learning to detect these two cognitive biases from Web logs of users’ actions. Our individual user level detection makes it truly personalized, and does not rely on annotated data. Instead, we start with two basic principles established in cognitive psychology, use modified training of an attention network, and interpret attention weights in a novel way according to those principles, to infer and distinguish between these two biases. The personalized approach allows detection for specific users who are susceptible to these biases when performing their tasks, and can help build awareness among them so as to undertake bias mitigation.
\end{abstract}





\maketitle

\section{Introduction}
Cognitive biases run through the disparate base literatures in fairness~\cite{jones2013perceptions}, in accountability~\cite{lerner1999accounting}, in transparency~\cite{lerner1999accounting,schaerer2018illusion}, in ethics~\cite{bostrom2006reversal}, in law~\cite{eskridge2001structuring,stroessner1996cognitive}, in medicine~\cite{croskerry2013mindless} and in discrimination~\cite{krieger1995content,tetlock2009implicit}. Biases pose in many forms~\cite{mehrabi2019survey}, with cognitive biases occupying key roles. Cognitive biases are mental shortcuts humans use in dealing with information and the environment, and which result in biased actions and behaviors (or, actions), unbeknownst to themselves~ \cite{tversky1974judgment}. A distinction of cognitive biases is that they occur without the human actor being aware of their impact on actions the person takes. That does not mean cognitive biases inflict less often than say, data bias and model / algorithm bias~\cite{amini2019uncovering,mehrabi2019survey}. On the contrary, the types of cognitive bias are many~\cite{pompian2011behavioral,haselton2015evolution} and they are omnipresent in a wide variety of situations with the potential for adverse effect on social good of the web, as the influential studies in the above literatures imply. Furthermore, these studies highlight the biased human actions that manifest from unobserved "mental" cognitive biases. With increased interest in using Machine Learning (ML) and Artificial Intelligence (AI) toward societal issues, the attention to detection of cognitive biases may be a worthwhile avenue to pursue. Our attempt is to offer an ML approach built on first principles from the cognitive psychology literature~\cite{hilbert2012toward}, for individual level detection of cognitive biases from actions, since actions are observed in data, but what is on the mind is not observable in data.  

It is accepted in Computer Science (CS) that ``Any remedy for bias must start with awareness that bias exists (page 54)"~\cite{baeza2018bias}. At the same time, ``Biases can be difficult to distinguish (page 2)"~\cite{white2013beliefs}. The major difficulty of cognitive biases is that, being mental shortcuts, they are not  observable, and thus cannot be directly ascertained. In CS, cognitive biases are recognized in  Information Retrieval (IR) research. Information search can become prey to unbeknownst cognitive bias of the user, if the information conforms to prior held beliefs and restricts the search~\cite{white2013beliefs,white2014content,white2015belief, pothirattanachaikul2020analyzing,gao2021addressing}. When users' actions are affected by cognitive biases then these biases can adversely impact tasks performed. Users can benefit from detection of their unbeknownst biases so that they can try to overcome those biases in their actions while performing their tasks~\cite{baeza2018bias}. The detection performed in IR is at the aggregate level of users with the help of annotated data. While the use of annotated data is received wisdom, any reasonably sized annotation of the many types of cognitive bias is not available, is costly and arguably, is not likely to be practicable. Instead, we start with first principles codified in cognitive psychology~\cite{hilbert2012toward} that are amenable to log data of users' actions, modify training of an ML network and interpret results according to those principles, to arrive at individual user level detection, without using annotated data. 

Although detection of cognitive biases can extend to many areas, the exposition of our approach uses the following context. For concreteness and as a running example for the rest of the paper, consider a Web based platform with which users interact to accomplish professional tasks; the interactions produce behavior log of users' actions. Examples include platforms used for searching information, selecting data, running analysis, and so on. Professional tasks are emphasized to distinguish from personal tasks users' perform on Web platforms because professional tasks require objectivity and cognitive biases become very important to examine~\cite{pompian2011behavioral}, whereas personal tasks can be guided by subjective personal beliefs and tastes. For such professional task oriented platforms, we focus on \textit{anchoring and adjustment bias} (hereafter, \textit{anchoring bias}) and \textit{recency bias}. The number of cognitive biases is large~\cite{pompian2011behavioral}; the scope of this paper confines to the detection of only two of them. With our focus on detection from observed actions we define these two biases as follows. Anchoring bias emanates from relying excessively on actions in \textit{initial} past periods to guide actions in current period, instead of making sufficient adjustment to learn from recent past. Recency bias emanates from relying excessively on actions in the \textit{recent} past periods to perform current period actions, instead of learning from earlier past. The research questions we ask are: 
\begin{itemize}
    \item How to distinguish between anchoring and recency biases and to uncover from users' behavior logs? 
    \item How to detect these biases at an individual level? 
    \item How to perform detection, when no annotation for biases is available?
\end{itemize}

We posit that these professional tasks come with objective goals and are \textit{meant} to be performed devoid of users' own personal taste, belief and preference. For example, when asked to seek information about market trends, a professional is expected to find \textit{facts} from reliable sources, rather than the information search becoming guided merely by one's personal beliefs. Another example is about fetching data for a presentation to senior executives, where a professional needs to get objective data, instead of data which has a personal slant. Yet another example is when analysts are required to use data, run models and produce findings that represent objectivity as much as is possible. In performing these kinds of tasks, users are likely to be inflicted with \textit{cognitive biases}~\cite{pompian2011behavioral}.

\iffalse
For concreteness, consider a digital platform with which users interact to accomplish professional tasks; the interactions produce behavior log of users' actions. Examples include platforms used for searching information, selecting data, running analysis, and so on. For such platforms, we focus on \textit{anchoring and adjustment bias} (hereafter, \textit{anchoring bias}) and \textit{recency bias}. Anchoring bias emanates from relying excessively on actions in \textit{initial} past periods to guide actions in current period, instead of making sufficient adjustment to learn from recent past. Recency bias emanates from relying excessively on actions in the \textit{recent} past periods to perform current period actions, instead of learning from earlier past. The research questions we ask are: 
\begin{itemize}
    \item How to distinguish between anchoring and recency biases and to uncover from users' behavior logs? 
    \item How to detect these biases at an individual level? 
    \item How to perform fully data driven detection, when no annotation for biases is available?
\end{itemize}
\fi

In CS, less attention is paid to cognitive bias detection, notable exceptions being~\cite{baeza2018bias,white2013beliefs, white2014belief,white2014content, white2015belief,pothirattanachaikul2020analyzing}. These works in IR examine the important problem of establishing biases in users' \textit{beliefs} and how that may impact search. In addressing our research questions around biases in users' \textit{actions}, we 
%attempt to add to the literature 
depart from the prior art on cognitive bias detection in three significant ways. First, unlike detection at an aggregate level, we propose detection for each individual user. Performed at the individual level, the detection helps specific users who are susceptible to these biases when performing their professional tasks. Alerting them conforms to the need for creating awareness in them about specific biases that may inflict their actions~\cite{baeza2018bias}.
Second, we analyze user logs in line with some previous work, but, unlike those, propose detection with an ML approach, by drawing from principles in cognitive bias literature~\cite{hilbert2012toward}.
Third, we do not use annotated data, which prior art relies upon. 

In particular, for a Web platform, the behavior logs of users' click-actions (hereafter, actions) are observable, although the mental states of the users remain unobservable. We posit that users' actions can be analyzed to ascertain cognitive biases. Drawing upon core principles from the cognitive bias literature~\cite{hilbert2012toward}, we introduce a principled formulation of the bias detection problem, which is amenable to log data. 
Modifying the training of a Hierarchical Attention Network (HAN)~\cite{attention:1}
%, tailored to our task and to log data, 
, we interpret attention weights from HAN according to the principles in a novel way to infer biases. 
Experiments find that our model can reliably detect these biases at the individual user level. As a first step toward individualized bias detection, our contributions are: 
\begin{itemize}
    \item Offering a new principled framework for identifying anchoring and recency biases in actions from behavior logs, without annotated (ground truth) data. 
    \item Proposing an ML model for individual level detection, which can overcome limitations of human studies and human annotations.
    \item Extending HAN through novel interpretation of its weights to this new domain of individual level bias detection.
\end{itemize}

We note that although our context is interactions with Web platforms for professional tasks, such interactions are de rigueur in personal tasks as well. Interactions in personal tasks include digital consumption such as music and movies, purchases across a broad swath of product categories, and communication exchanges on social sites, among others. For an individual user, these interactions are often guided by her own personal taste and preference. While these actions could also be inflicted by cognitive biases of the individuals as well as biases of the algorithms that the platforms rely upon (e.g., recommendations), examination of those are outside the scope of this paper.

\section{Related Literature}
Impact of the Web and related technologies on society can be observed through multiple lenses of fairness, accountability, transparency, ethics, law, medicine, discrimination and others. Examination of each lens through its literature can contribute to a deeper understanding of what comprises social good, and calls out the role of biases. In particular, we note the connection to cognitive biases recognized across literatures in fairness~\cite{jones2013perceptions}, in accountability~\cite{lerner1999accounting}, in transparency~\cite{lerner1999accounting,schaerer2018illusion}, in ethics~\cite{bostrom2006reversal}, in law~\cite{eskridge2001structuring,stroessner1996cognitive}, in medicine~\cite{croskerry2013mindless} and in discrimination~\cite{krieger1995content,tetlock2009implicit}. For example, in medicine, “Over the past 40 years, work by cognitive psychologists and others has pointed to the human mind’s vulnerability to cognitive biases, logical fallacies, false assumptions, and other reasoning failures. More than 100 biases affecting clinical decision making have been described, and many medical disciplines now acknowledge their pervasive influence on our thinking"~\cite{croskerry2013mindless}. Or, with regard to ethics, it is stated that "Recognizing and removing a powerful bias will
sometimes do more to improve our judgments than accumulating or
analyzing a large body of particular facts. In this way, applied ethics
could benefit from incorporating more empirical information from psychology
and the social sciences about common human biases"~\cite{bostrom2006reversal}. In accountability research an influential review paper~\cite{lerner1999accounting} highlights the key position of cognitive bias in "accounting for the effects of accountability". In particular they propose and find evidence that "the effect
of accountability depends on a complex host of moderators, including
the cause of a given bias, the type of accountability, and
the decision maker's knowledge of formal decision rules [pp.259]"~\cite{lerner1999accounting}.

Research on cognitive biases, with roots in Psychology~\cite{edwards1954theory, simon1955behavioral,simon1956rational, haselton2015evolution} in 1950s, permeates many disciplines. Prior art on cognitive biases come from Cognitive Psychology \cite{haselton2015evolution}, Psychology \cite{hilbert2012toward, tversky1974judgment}, Behavioral Finance \cite{barberis2003survey,pompian2011behavioral}, Economics \cite{rabin1998psychology,rabin1999first}, among other fields. 
``People rely on a limited number of heuristic principles which reduce the complex tasks of assessing probabilities and predicting values to simpler judgmental operations (page 1124)"~\citet{tversky1974judgment}. Focusing on judgmental operations we restrict to actions humans take because actions are manifestations of cognitive biases. As evidence, research in Behavioral Finance scrutinizes how cognitive biases manifest in choice of actions (decisions) of financial analysts \cite{pompian2011behavioral,barberis2003survey}. It is established that even actions with large financial stakes may fall prey to one of a long list of cognitive biases, unbeknownst to them \cite{pompian2011behavioral}. In particular, we focus on Anchoring bias and Recency bias, as a first step toward individualized detection. The rich prior art in these disciplines motivates our thinking. These works rely on human studies - both laboratory setting and crowd-sourcing platforms (e.g., MTurk) - where vignettes of information-scenarios are presented to participants, based on which they choose among actions~\cite{barberis2003survey}. But, human studies suffer from four deficiencies: (i) limited to a few users, (ii) use scenarios instead of real data, (iii) detection is at an aggregate level, and (iv) not data driven. We overcome all four deficiencies.

In computer science, study of cognitive biases is relatively recent, although position bias and domain bias are examined for some time ~\cite{joachims2007evaluating,fortunato2006topical,ieong2012domain}. Pioneering work within IR ~\cite{white2013beliefs,white2014belief} examine beliefs and, anchoring and confirmation biases in the context of search. In search for medical and health information, it is shown that users' pre-search beliefs bias search results and reinforces human bias~\cite{white2014content,white2015belief,pothirattanachaikul2020analyzing}. 
Besides showing existence of biases in search, the research extends to how beliefs change as search progresses~\cite{white2014belief,white2015belief}. The works show how belief dynamics are affected by search results. Collectively, these works use multiple methods including survey, human annotations, human experimental studies, and analysis of large search engine logs to detect bias at the \textit{aggregate} level.  
The use of search log data is reduced to aggregate metrics, but are not used in modeling the sequence of actions users take. The aggregate metric is used as a basis of evaluation, and bias identification is shown for a group of individuals. By not modeling the sequence of actions the prior art forgoes an opportunity to learn from them. To our best knowledge, \textit{individual} level bias detection from user behavior logs is novel, distinguishing our work. Moreover, we model the sequence of actions using ML to learn biases. 
A conceptual paper \cite{kliegr2018review} describes the potential effects of cognitive biases on interpretation in ML and draws attention to the importance of studying cognitive biases, lending support to our endeavor, although it does not offer a model to uncover them.
A large body of ML research is devoted to biases in data, models and algorithms ~\cite{jordan2015machine,amini2019uncovering,mehrabi2019survey}, 
but does not address detection of cognitive biases. 

\section{The Principles}
Individualized detection of cognitive biases from behavior logs has no established framework. We do not have ground truth for anchoring and recency biases in actions. Our principled framework builds upon two principles established in the literature on cognitive biases. 

\textit{Principle PN1}: A cognitive bias is a systematic, predictable \textit{deviation} from a \textit{norm}~\cite{hilbert2012toward}. 

\textit{Principle PN2}: To be classified a bias, the deviation has to occur \textit{consistently} and cannot be random~\cite{hilbert2012toward}. 
 
Following principle \textit{PN1}, we compute a norm for actions, and then judge individual's bias by examining deviation of his / her actions from the norm. Adhering to principle \textit{PN2}, we test for consistency of deviation for each user across many instances, to rule out the alternative of randomly occurring deviations. Unlike our personalized detection, in prior art, aggregate level deviations are shown with group of participants by examining a predictable pattern of deviations either from a norm defined by a utility model or a probability model~\cite{hilbert2012toward}, or, from a norm of annotated truth~\cite{white2014belief}. \textit{Predictable deviation} conforms to \textit{PN1}. But, to satisfy \textit{PN2} requires examination of \textit{consistency of deviation over time}, which is not embedded in the approach of this prior art. As a deviation from this prior art, our approach starts from both \textit{PN1} and \textit{PN2}, and then builds the ML approach to conform to them. 

\section{Data and Framework}\label{section_conceptual_framework}
We now describe the data, present the framework, provide rationale of the model, and posit the properties of deviation for bias detection. The mathematical formulation of deviation is presented in the sub-section on Training Strategy, equation~\ref{eq:def_dev}.  

\subsection{Data and Preprocessing}\label{sub-section-type_of_data}
The dataset comes from a Web enabled analysis platform, where users are marketing analysts (hereafter, \textit{users}). Users interact with the UI to fetch data stored in the platform's database and to perform analysis tasks (hereafter, \textit{task}) with the help of tools available on the UI. Behavior logs show time-stamped, click actions (actions) of users, for different tasks. Each user may perform multiple tasks; some tasks are performed multiple times a week, while others are performed once a week. Each user's visits to the UI across time are stitched. A \textit{visit} comprises a sequence of actions for a task. Examples of actions include \textit{new report, select data, select model,} etc. From the data of over two months, for nearly 1,050 users, 81 click-actions covering 95\%-ile of all actions taken by users, are grouped into 33 action-categories (hereafter, \textit{actions}), based on similarity of click-actions. This grouping is done for model parsimony and is verified by experienced analysts. The dataset uses these 33 actions. We drop users with nine or fewer visits, as they present few data points for activities. To avoid outliers, we drop users with 83 or more visits $(mean+3*SD)$. After preprocessing, number of users for bias detection is 244 and the number of visits per user varies from 10 to 82. Remaining users, numbering around 800, contribute to establishing the norm (principle \textit{PN1}), described in Subsection Training Strategy. 

\subsection{Conceptual Framework}\label{sub-section-framework}
For context, the framework is described with respect to the data at hand. There is nothing unique about this type of data. Increasingly Web enabled platforms are being used for completing professional tasks as the move to the cloud expands, making the availability of this kind of behavior log data of users' actions quite common. Notably, the framework and the modeling approach naturally extends to other data with behavior log of actions, as applied to professional tasks, which require objectivity. We assume that different tasks require different sets and sequences of actions to be performed. For example, if the analysis task is \textit{segmentation} of consumers, the set of actions is \{\textit{select timeline for fetching data, select attributes, set rules for each attribute, segment creation, etc}\}. However, for a different task, \textit{attribution analysis}, the set of actions is \{\textit{select target label, select duration of analysis, run attribution analysis, etc}\}. 

It is crucial to fix a task since selection of actions is tied to the task, and detection of biases in actions is defined for the task. 
Given a specific task, our goal is to learn, for each user, over many visits, how sequences of actions performed in past visits impact choice of sequence of actions in the current visit. The model learns the degree of influence each of the past visits has on the current visit. A relevant question is that if a task requires a user to perform a set of actions in a sequence, pre-defined by the platform's UI, does that constitute any form of bias? The answer is \textit{no}. To address this question, our modeling approach recognizes that for a task performed on the UI, some actions are taken following a common sequence inherently required by the task, while other sequence of actions is individual specific. 
In line with principle \textit{PN1}, we compare individual user's actions, over visits, against the \textit{norm} of common sequence, to detect biases in individual user’s choice of actions. 
Following principle \textit{PN2}, to check \textit{consistency} of deviations, we analyze a large number of visits for each user. To test sensitivity of results, we experiment with different numbers of visits.

\begin{figure}[t]
    \centering
    \includegraphics[width=1.0\linewidth]{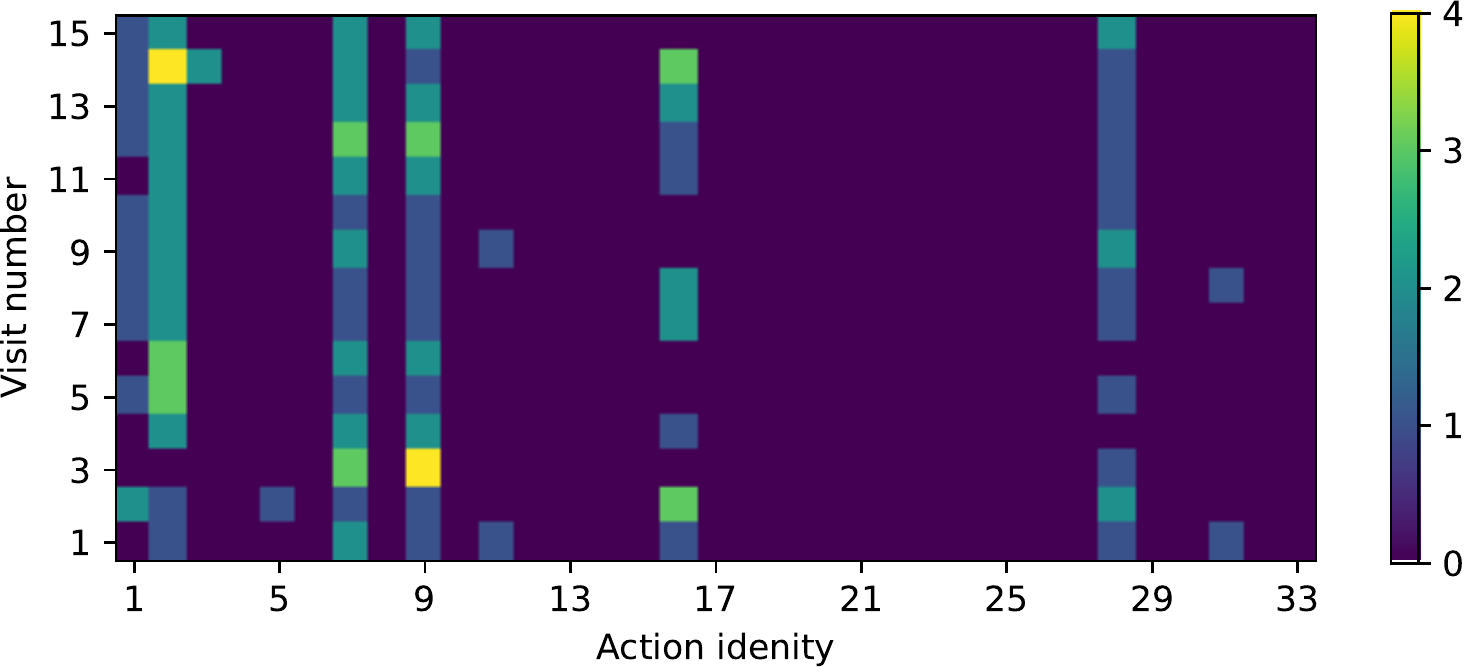}
    \caption{Heatmap of 33 actions performed by a user over 15 visits, visualizing frequency of each action across visits.}
    \label{anchoring_bias}
\end{figure}

\subsection{Rationale for our Model}
\label{section_intuition_rationale}
Without annotated (ground truth) data we cannot model detection of these biases in a direct manner. Instead, we take an indirect approach, whose intuition is offered through a simple visualization. Later we present a formal model. 

The heat map in Figure~\ref{anchoring_bias} shows actions of a randomly chosen user, for a specific analysis task done periodically. We observe the frequency of each action, out of 33 possible actions for the task, over 15 different visits, with the visits spread across weeks. The frequencies depicted by shades of color show that her choice of actions in later visits (visit numbers 10 - 14) is rooted in choice of actions in initial visits (visit numbers 0 - 5). There is little exploration of new actions in visit number 6 and thereafter. The user is likely anchored to actions in early visits and does not make much adjustment during later visits, possibly indicating anchoring bias in actions. This intuitive rationale is formalized in a model.  
We need a model, applicable to all users' sequence of actions, which 
(a) assigns weights across past visits, such that weights indicate degrees of dependency of actions in current visit upon actions in past visits; and (b) distinguishes between anchoring and recency biases.  
Task specific requirements of an UI necessitate users to mimic certain action-sequence over repetitive visits and does not constitute a bias. Based on principle \textit{PN1}, our approach should compute the norm (common effect) for the task and then identify each individual's bias as \textit{deviation} from the norm.

\subsection{Properties of Deviation for Bias Detection} Above description of anchoring and recency biases maps to their detection as follows: 

(1) For \textit{anchoring bias} the deviations have \textit{larger weights} for early visits relative to recent visits. 

(2) For \textit{recency bias} the deviations have \textit{smaller weights} for early visits relative to recent visits.  

(3) When the deviations have \textit{similar weights} for early and later visits, it shows \textit{lack of evidence} of either bias. Note that for (3), we cannot claim the user is cognitively unbiased. 

To examine whether these properties conform to the notion of these two biases in the minds of experts, we recruited 9 experienced marketing analysts, with median number of years of experience as 7 years. While the number of study participants is small, we note that their expertise and experience add much value to the study since unlike inexperienced participants these experts can relate to the study-context well. 
No input from experts enter the model or analysis. These experts gave their interpretation of the patterns of deviation. To every expert three separate graphs were shown, each constructed from \figurename ~\ref{relative_attn}. Each graph showed three lines: graph 1 for flat pattern, graph 2 for decreasing pattern, and graph 3 for increasing pattern. For a deeper exposition of the task to these experts, we also provided them with more context by showing them several tables of sequences of actions from behavior logs of marketing analysts, across many visits. With each graph shown separately, experts responded by selecting one option out of four possible options: (i) neither anchoring nor recency, (ii) anchoring, (iii) recency, (iv) cannot tell. Increasing weights were associated with recency bias by 8 out of 9 experts, confirming our interpretation for recency bias. Out of 9 experts, decreasing weights were associated with anchoring bias by the modal frequency of 4 analysts, while 1 analyst selected (i), 2 selected (iv), and 2 did not answer. This provides support for our interpretation of anchoring bias. Overall, these experts corroborate the interpretation of pattern of deviations for bias detection. 

\begin{figure}[t]
\centering
    \includegraphics[width=0.95\linewidth]{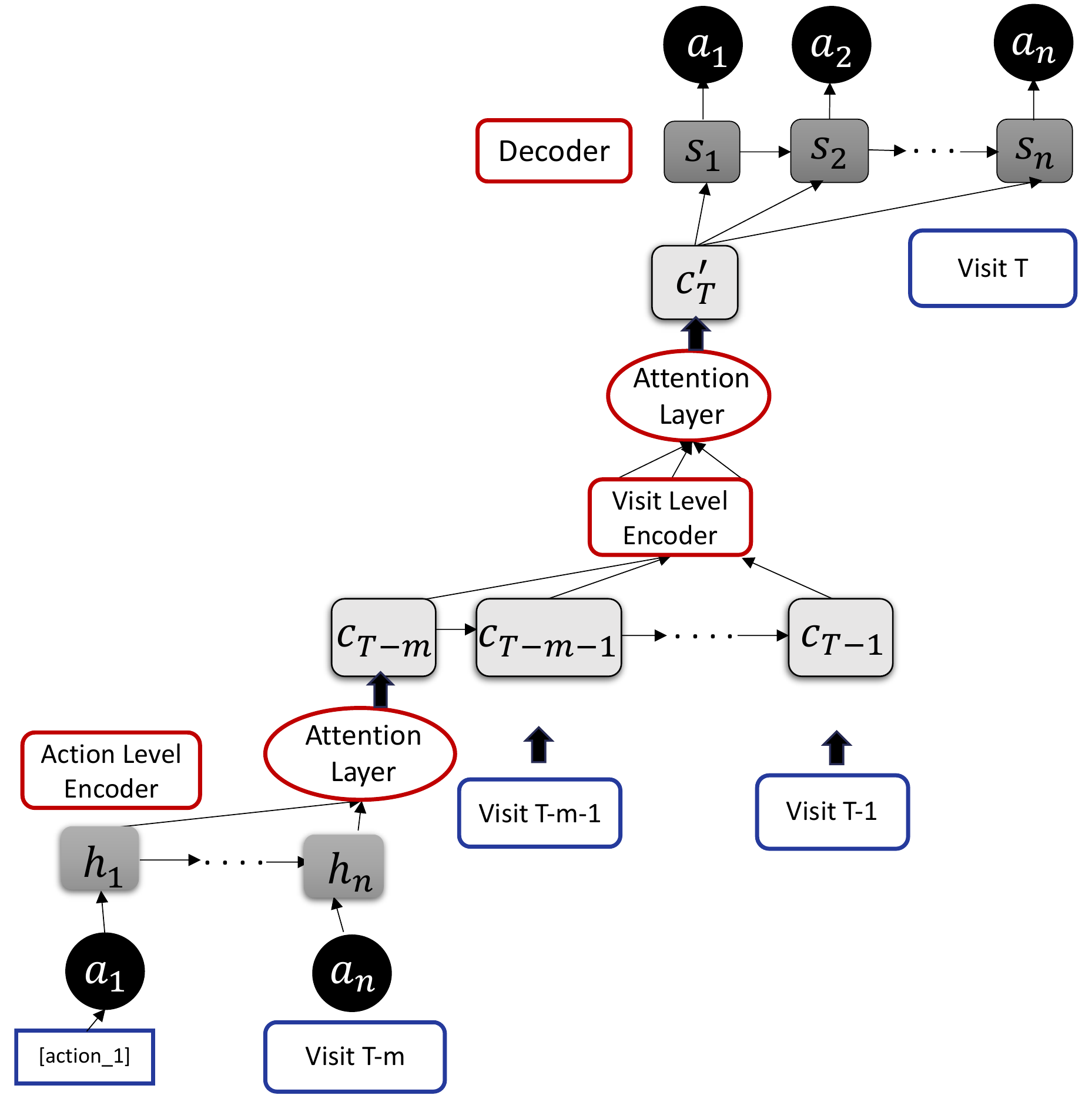}
    \caption{Hierarchical Attention Model with two encoders, corresponding to two levels, and one decoder. In bottom (action) level, sequence of actions $a_1$,..,$a_n$ in one visit, is encoded to produce visit context vector; in middle (visit) level, context vectors for a sequence of $m$ visits ($T-m$),..,($T-1$), within a window, are encoded to yield a window specific context vector with weights for each visit. The decoder predicts sequence of actions for visit $T$.}
    \label{network_diag_2}
\end{figure}

\section{Model - Hierarchical Attention Network}\label{section_HAN}
To check for properties of deviation stated above, we estimate weights for visits by a formal model. To reiterate, on the analysis UI, the task is performed through a sequence of actions. 
Since actions are selected within a visit, and bias is to be identified by relating action-sequences in several past visits to action-sequence in current visit, we employ a Hierarchical Attention Network (HAN)~\cite{attention:1} with \textit{action level attention} and \textit{visit level attention}. The gist of the modeling framework is shown in \figurename ~\ref{network_diag_2}. The HAN comprises dual encoders and a decoder, with attention layers at each encoder~\cite{attention:1,DBLP:journals/corr/WangWTOCL16, hierarchical_attn:1}. 
 
We employ the visit-level attention weights as the degrees of influence of action-sequences in %each of 
the previous visits
%$T-m,\ldots,T-1$ 
on the action-sequence in the current visit $T$. These weights are in turn learned by sequential prediction of actions in visit $i=T$, given the sequence of actions of previous $m$ visits, i.e. $i=\{T-m,\ldots,T-1\}$, with each visit comprising $n$ actions. Say $a_{i,1}, a_{i,2}, \ldots, a_{i,n}$ represents the action sequence in the $i$-th visit. 
In the action-level encoder, the LSTM processes this action input sequentially to output hidden vector $h^a_{i,j}$, for $j=1,2,\ldots,n$. The action-level context vector $c_i$ captures the complete $i$-th action sequence by employing an attention mechanism on top of these hidden representations. The context vector $c_i$ is the weighted average of $h^a_{i,j}$, where \textit{attention} weights $\alpha_{ij}$ represent importance of $j$-th action in the context of $i$-th visit. More formally,
\begin{equation}\label{eq:action-context}
    c_i = \sum_{j=1}^n \alpha_{ij} h^a_{i,j}.
\end{equation}
Similarly, visit-level encoder estimates state representation of a sequence of $m$ successive visits, $T-m$ to $T-1$, using an LSTM. The input to this encoder are context vectors $c_{T-m},\ldots,c_{T-1}$ calculated using equation \ref{eq:action-context} which encodes information from actions performed in each of these visits. The visit-level LSTM calculates the hidden representations $h_i^v$ for $i=T-m, \ldots, T-1$ and the visit-level context vector is the weighted average of hidden states $h_i^v$ of this LSTM. We represent this visit-level context vector of the visits $i=T-m, \ldots, T-1$ as $c'_T$, as it used to predict the action-sequence in the subsequent visit (i.e. $i=T$). Thus,
\begin{equation}
    c'_T = \sum_{i=T-m}^{T-1} \alpha'_{i} h^v_{i}.
\end{equation}
This attention mechanism allows the decoding process to appropriately attend to different visit sequences during decoding, emulating the action selection process of a user.
The sequential decoder uses context vector $c_T'$ and predicts sequence of actions in visit $T$. The $j$-th term in output sequence is computed as softmax of the $j$-th hidden state $s_j$ of decoder level LSTM. 
The decoder assigns the probability of $j$-th action, $p(a_{T,j})$ conditional on the context vector $c_T'$ and its previous hidden state $s_{j-1}$, over the entire action space. Thus, the probability $p(a_{T,j})$ is conditional on the actions $1:n$ from visits $T-m,\ldots, T-1$ and the actions $1:j-1$ from the visit $T$. 
The end-to-end model is trained to minimize the Cross-Entropy loss of predicting the action-sequence $\{a_j\}_{j=1}^n$ of $T$-th visit from action-sequence of sequence of visits $T-m,\ldots,T-1$, across rolling windows of visits. The overall training loss is defined as,
\begin{equation}
    \mathcal{L} = \underset{T\sim D}{\mathbb{E}} \big[\sum_{j=1}^n - a_{T,j}.p(a_{T,j}|a_{T-m, 1:n}, \ldots, a_{T-1,1:n}, a_{T,1:j-1} )\big].
\label{eq:training_loss}
\end{equation}
Here, $D$ is the training corpus from which the window ($T-m,\ldots,T-1$) of visits is sampled, as described next.

\subsection{Training Strategy}\label{subsection_training_details} 
Following principle \textit{PN1}, a norm is computed through \textit{Common training}, which represents actions required by the UI to be commonly taken by all users for the \textit{CS} task. This norm is a set of attention weights (hereafter, weights) over visits, common for all users. Then, \textit{Personalized training}~\cite{10.1145/2556195.2556234} is performed for each user. The personalized training is used to establish deviations in actions of individual user from the norm. For each user, the weights over visits are computed, and then deviations in weights from that of the norm are computed. Following principle \textit{PN2}, an \textit{Extended-personalized training} yields robust weights for each individual user, which tests whether the deviations in weights observed for each user, are consistent over rolling windows. 

\begin{figure}[t]
    \centering
    \includegraphics[width=\linewidth]{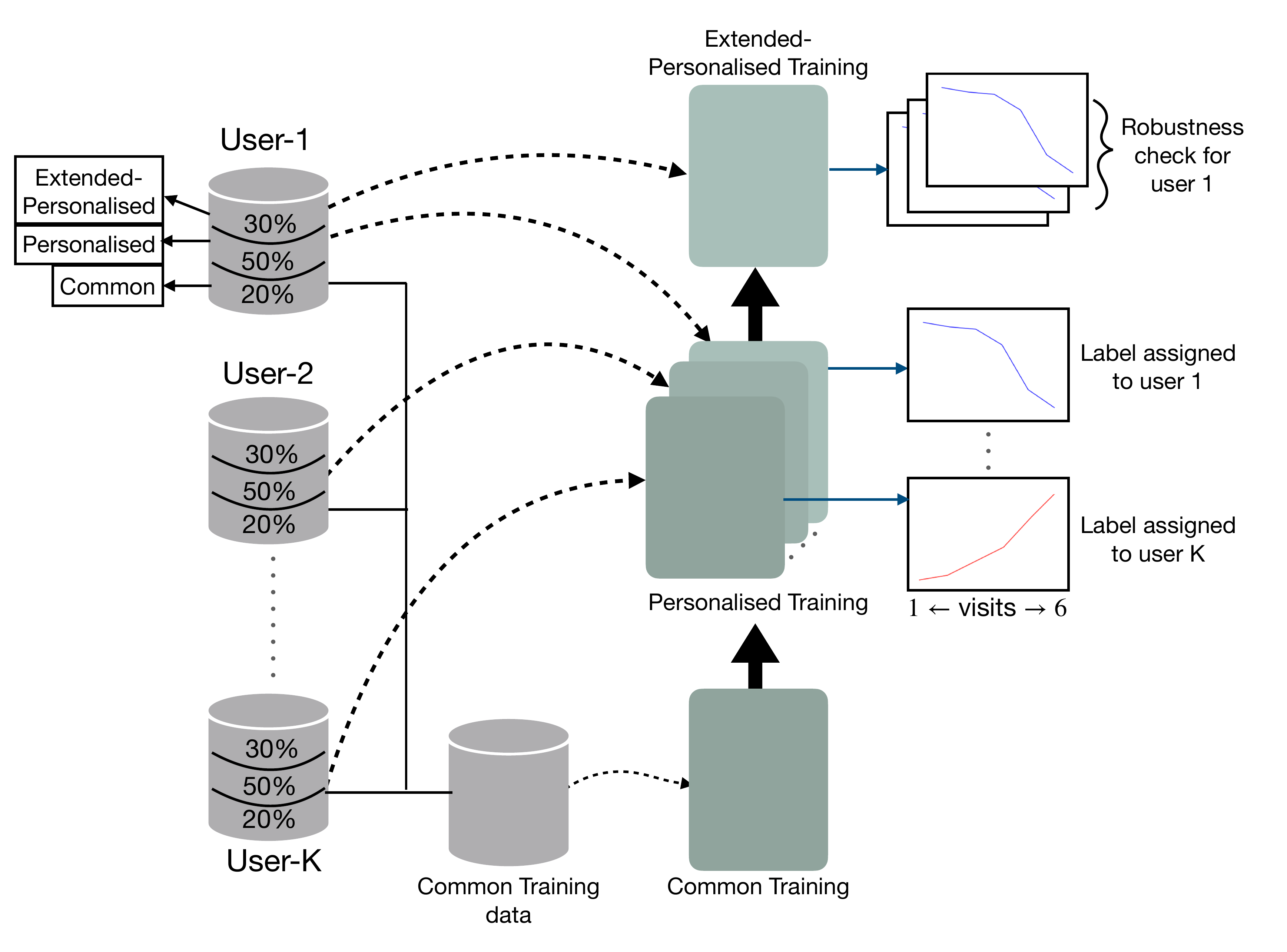}
    \caption{Training strategy for personalized Bias Detection and for Reliability.  The 20:50:30 data split, for each of K=244 users, (Left section) contributes respectively, to the three successive training steps \textit{Common}, \textit{personalized}, \textit{Extended-personalized} (Middle section). personalized training step yields attention weights, over 6 visits, for each user (Rightmost Section). The graphs show pattern of deviation in weights from the norm for each user, and are used to assign label for types of bias. Reliability is assessed in the step, Extended-personalized training. Topmost group of graphs yields weights over 6 visits for several rolling windows of visits, for each user. Reliability is tested across rolling windows.}
    \label{training_diag}
\end{figure}

The training strategy in \figurename~\ref{training_diag} has the left section showing that each of 244 users' data are split 20:50:30 to contribute respectively, for common : personalized : extended-personalized training. The 20\% data from each of 244 users and data from the remaining users, numbering around 800, form input for common training. The middle section shows the three successive training steps of \textit{Common}, leading to \textit{personalized}, then followed by \textit{Extended-personalized} training. The rightmost section shows 
graphs of \textit{deviation} in personalized training attention weights from that of common training weights (norm), over 6 visits, for each of K=244 users. We define deviation in weights as, 
\begin{equation}
    \delta_{k,i} = \alpha'_{k,i} - \alpha'_{i},
    \label{eq:def_dev}
\end{equation}
where, $k$ is user, $i$ is visit sequence number in a rolling window; $\delta_{k,i}$ and $\alpha'_{k,i}$ are, respectively, deviation, and personalized attention weight of user $k$ for visit $i$; and $\alpha'_{i}$ is common (across users) attention weight for visit $i$.  

The pattern of deviations for each user results in assigning label for types of bias. Moreover, Reliability in pattern of deviations, for each user, is assessed in the additional step, namely the extended-personalized training. The topmost part of the rightmost section shows that a set of deviations over 6 visits emerges, for each rolling window of visits. Since each user has several rolling windows of visits, several sets of deviations emerge. The \textit{change} in pattern of deviations over visits, \textit{across} rolling windows, can be enumerated, for every user. For a user, if the pattern of deviations \textit{changes across rolling windows} then the consistency of individual-user-specific bias detection, as required by principle \textit{PN2}, can be called into question. The Reliability test is performed for all K=244 users.

\section{Experimental Details}\label{section:experiments}
\textbf{Training Data.} Our dataset restricts to the most popular task on this data-provider's UI, \textit{segmentation analysis}. As a first study in ML based individualized detection, we showcase our approach using this task as an exemplar. The approach extends to other analysis tasks generating behavior log data. Our detection model is implemented for this segmentation analysis task, identified in logs through the focal action, \textit{Creating Segment} (hereafter, \textit{CS} or \textit{task}). A \textit{sequence} of additional actions is taken around the focal-action - both \textit{before} and \textit{after} the focal-action - to capture the context of the focal-action. 
We use a sequence of 10 actions immediately before \textit{CS} and 10 actions immediately after \textit{CS}, yielding $n$=21 actions as input, per visit. This action set is labeled (\textit{CS -/+ 10}). The choice of 10 is informed by histogram of types of action per visit, and the number of actions and number of visits available in data for each user. 
The number of visits for each user is broken into \textit{rolling windows} (hereafter, windows) of successive visits, satisfying two opposing forces: (i) each window spans a minimum number of visits so that anchoring bias can be uncovered going back to somewhat-past visits; and (ii) an adequate number of windows for training HAN. With $k$-th user's data having altogether $N_k$ visits and $m$ successive visits in a window, we obtain $N_k-m$ windows of visits for $k$-th user. For example, if a user has 20 visits, $m=6$ implies 14 windows. Action-sequences in visits $1$-$6$ predict action-sequence in visit $7$, ..., those in visits $14$-$19$ predict for visit $20$. Across users, the number of windows varies since the number of visits range from 10 to 82. Table~\ref{tab:data-size} shows statistics for number of visits by data split. 
First we present results with $m=6$. Then, we conduct sensitivity experiments by varying the number of visits in the window. A window of $m$ visits is one data point for training. 

\textbf{Hyper-parameters.} Action-level encoder is a bi-directional LSTM to capture the context of the action-sequence. The hidden space dimensions for action-level encoder, visit-level encoder, and decoder LSTMs are $512$, $2048$ and $2048$, respectively. Randomly initialized parameters are trained with loss in equation \ref{eq:training_loss}, for over $25$ epochs, until convergence. An Adam optimizer with learning rate $10^{-3}$ and batch size $40$ is used. Dropout to the output of each LSTM layer occurs with $p_{drop}=0.2$.
\iffalse
\begin{figure}[t]
    \centering
    \includegraphics[width=0.8\linewidth]{Images/pers_weights_relative-2.pdf}
    \caption{Personalized Bias Detection. Using personalized training, deviation $\delta_{k,i}$, over 6 visits, for 244 users is shown. Trends in $\delta_{k,i}$ are: decreasing (blue) for anchoring, increasing (red) for recency, flat (greenish) for inconclusive.}
    \label{relative_attn}
\end{figure}
\fi

\begin{table}[t]
    \centering
    \caption{Statistics for Number of Visits}
    \label{tab:data-size}
    \begin{tabular}{c|c|c}
        \hline
        \textbf{Data Split} & \textbf{Mean} & \textbf{SD} \\
        \hline
        Common & 8.45 & 3.47 \\
        Personalized & 12.60 & 5.01\\
        Extended-personalized & 8.62 & 3.09\\
        \hline
    \end{tabular}
\end{table}

\begin{table}[t]
    \centering
    \caption{Frequency distribution based detection. For each number of successive visits, quantity and percentage of statistically inconclusive detection, out of all 244 users, are shown. We find more than 2/3-rd inconclusive detection.}
    \label{tbl-freq-actions}
    \begin{tabular}{l|c|c}
        \hline
        \textbf{No. of successive Visits} & \textbf{Inconclusive} & \textbf{Percentage} \\
        \hline
        $m=4$ & 172 & 70.5 \\
        $m=6$ & 167 & 68.4 \\
        $m=8$ & 176 & 72.1 \\
        \hline
    \end{tabular}
\end{table}

\begin{figure}[t]
    \centering
    \includegraphics[width=0.8\linewidth]{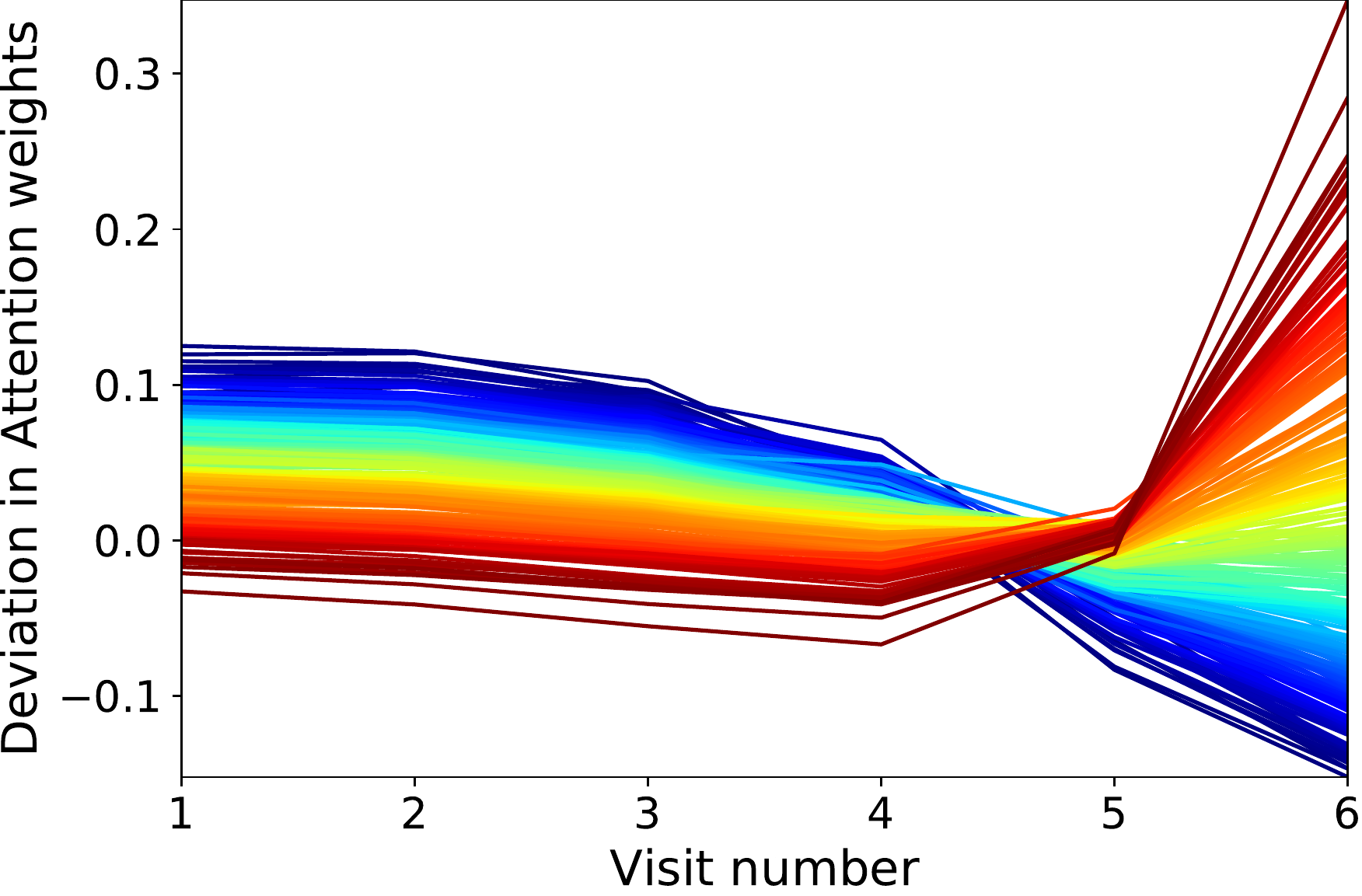}
    \caption{Personalized Bias Detection. Using personalized training, deviation $\delta_{k,i}$, over 6 visits, for 244 users is shown. Trends in $\delta_{k,i}$ are: decreasing (blue) for anchoring, increasing (red) for recency, flat (greenish) for inconclusive.}
    \label{relative_attn}
\end{figure}

\section{Results}\label{section_results}
\subsection{Baseline}
In the absence of established baseline for personalized bias detection, we construct a baseline from the frequency distribution of actions to detect these biases. The idea is to determine whether we can identify individual biases depending upon the repetitiveness / similarity of the sets of actions that users employ over different visits. For each user, each visit yields a frequency distribution over actions. We use the Wasserstein distance\footnote{Implemented per SciPy library (https://www.scipy.org)} $w(\nu_i, \nu_j)$ between frequency distributions ($\nu$) of actions across visits $i$ and $j$ as a measure of similarity in sets of actions between visits \cite{Ramdas_2017}. For discrete frequency distributions $\nu_i$ and $\nu_j$, the Wasserstein distance is defined as 
\begin{equation}
    w(\nu_i, \nu_j)=\sum|U_{i}-U_{j}|,
\end{equation}
where $U_{i}$ is the cumulative distribution function of frequency distribution $\nu_i$.
We also need to distinguish between anchoring and recency biases. Consider successive visits. If we focus merely on two successive visits, both anchoring and recency biases provide identical similarities between successive visits, because anchoring bias suggests that visit 1's pattern of actions carries over to visit 2, and recency bias suggests that visit 2's pattern of actions is rooted in those of visit 1. Thus, similarity between two successive visits is not informative in distinguishing anchoring bias from recency bias. It behooves looking for similarities between visits that are farther apart. Now consider a set of $m$ successive visits, where, $m>>1$. We compute $m$ Wasserstein distances in frequency distributions as follows: Distance between visit 1 and visit $m+1$, between visit 2 and visit $m+1$, ..., between visit $m$ and visit $m+1$. We compute the slope across these $m$ distances. If the slope is significantly decreasing (increasing), we identify with anchoring (recency) bias. That is, we expect in anchoring (recency) bias, the dependencies of visit $m+1$ on visits $1,2,\ldots,m$ progressively decreases (increases). Performing this analysis we find that a large number of inconclusive detection results. As shown in Table~\ref{tbl-freq-actions}, with $m$ = $4$, $m$ = $6$, $m$ = $8$, we get that 70.5\%, 68.4\%, 72.1\%, respectively, of all users cannot be classified into either of these two biases. It is to be noted that the failure to classify a user into either an anchoring bias or a recency bias does not imply that the user is cognitively unbiased. This baseline may be failing to detect biases of some users. 

\subsection{Model results - Personalized Bias Detection}
All detection results are shown for deviation in weights, $\delta_{k,i}$, defined in equation~\ref{eq:def_dev}. 
%\subsection{Personalized Bias Detection}\label{subsection_experiment_1}
\figurename ~\ref{relative_attn} shows deviations, $\delta_{k,i}$, $i$=1,..,6 visits, for each user $k$, where $\alpha'_{k,i}$ and $\alpha'_{i}$ are, respectively, outputs of the last rolling window of personalized training and common training. Across $244$ users, $\delta_{k,i}$ values trend from increasing (red shades) to flat (greenish) to decreasing (blue). Thus, there are substantial differences across users in tendencies to rely on actions in visits 1 and 2, \textit{versus} actions in visits 5 and 6, to guide actions in visit 7. Following the Properties of Deviations, the increasing trend (red) indicates recency bias since actions in visit 7 are influenced more by actions in visits 5 and 6, instead of by actions in visits 1 and 2. The decreasing trend (blue) indicates anchoring bias since visits 1 and 2 have higher influence, relative to visits 5 and 6, on actions in visit 7. The lines with almost equal deviation (greenish), indicate lack of evidence for any of these two biases. Notably, more users in blue shades than in red shades indicate higher incidence of anchoring bias relative to recency bias.

\begin{figure}[t]
\centering
  \includegraphics[width=.75\linewidth]{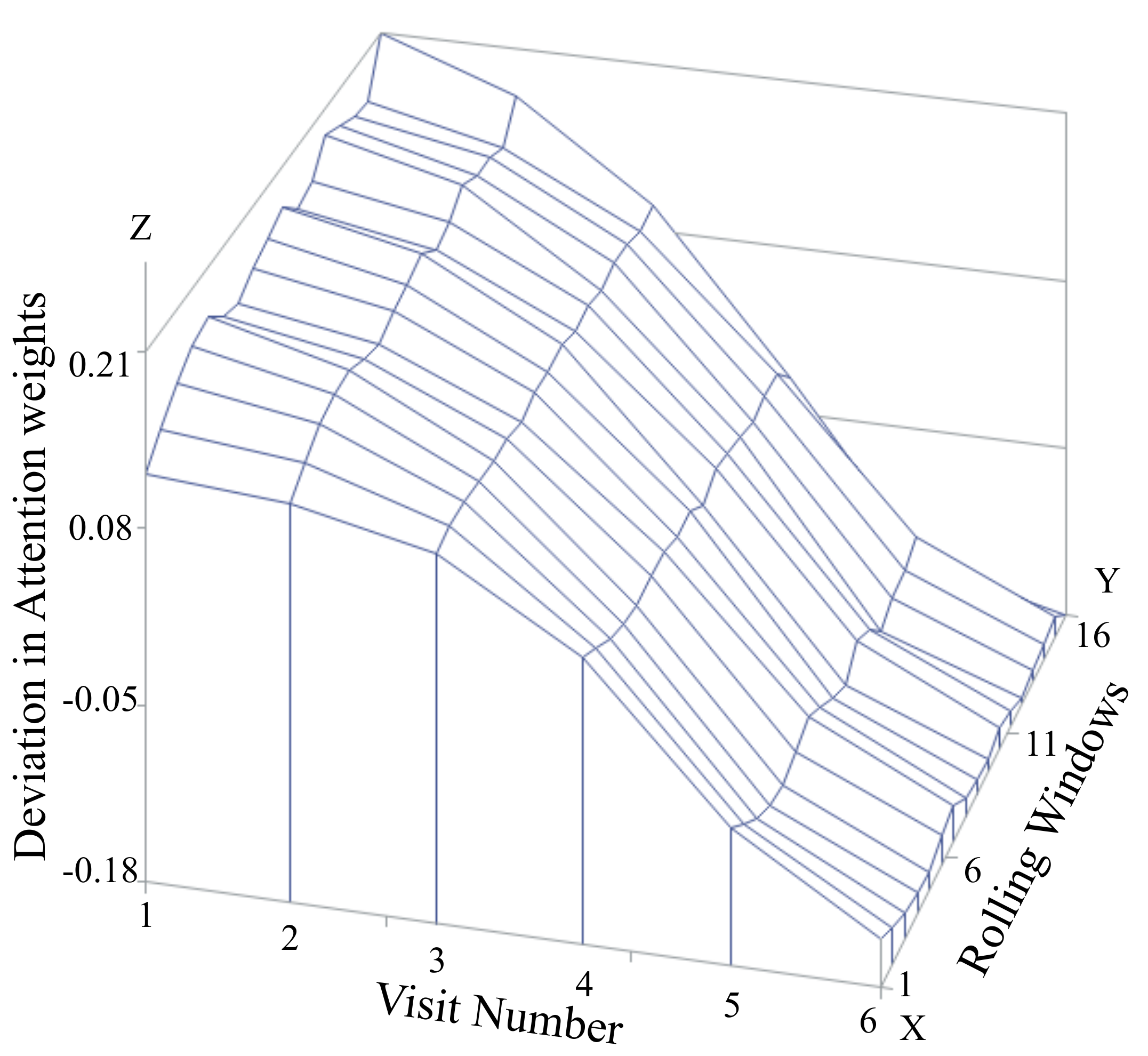}
  \caption{Detection-Reliability. For a user's Extended personalized training, deviation $\delta_{k,i}$ values (Z-axis) for 6 visits (X-axis), across 17 rolling windows (Y-axis) show \textit{consistency} in decreasing trend of $\delta_{k,i}$ over visits across windows, implying reliability of anchoring bias detection.}
  \label{attn_wt_3d}
\end{figure}

\begin{table}[t]
    \centering
    \caption{Detection-Reliability. Distribution of reliable slope coefficients, $\beta_{k,1}$, for deviation $\delta_{k,i}$ over 6-step visit sequence, among 244 users, using statistical test at $\alpha$ = 0.05.}
    \label{tbl-p-value}
    \begin{tabular}{l|c|c}
        \hline
        \textbf{Bias} & \textbf{Frequency} & \textbf{Percentage} \\
        \hline
        Anchoring & 192 & 79 \\
        Recency & 29 & 12 \\
        Inconclusive & 23 & 9 \\
        \hline
    \end{tabular}
\end{table}

\begin{figure*}[t]
     \centering
     \begin{subfigure}
     \centering
     \includegraphics[width=0.27\textwidth]{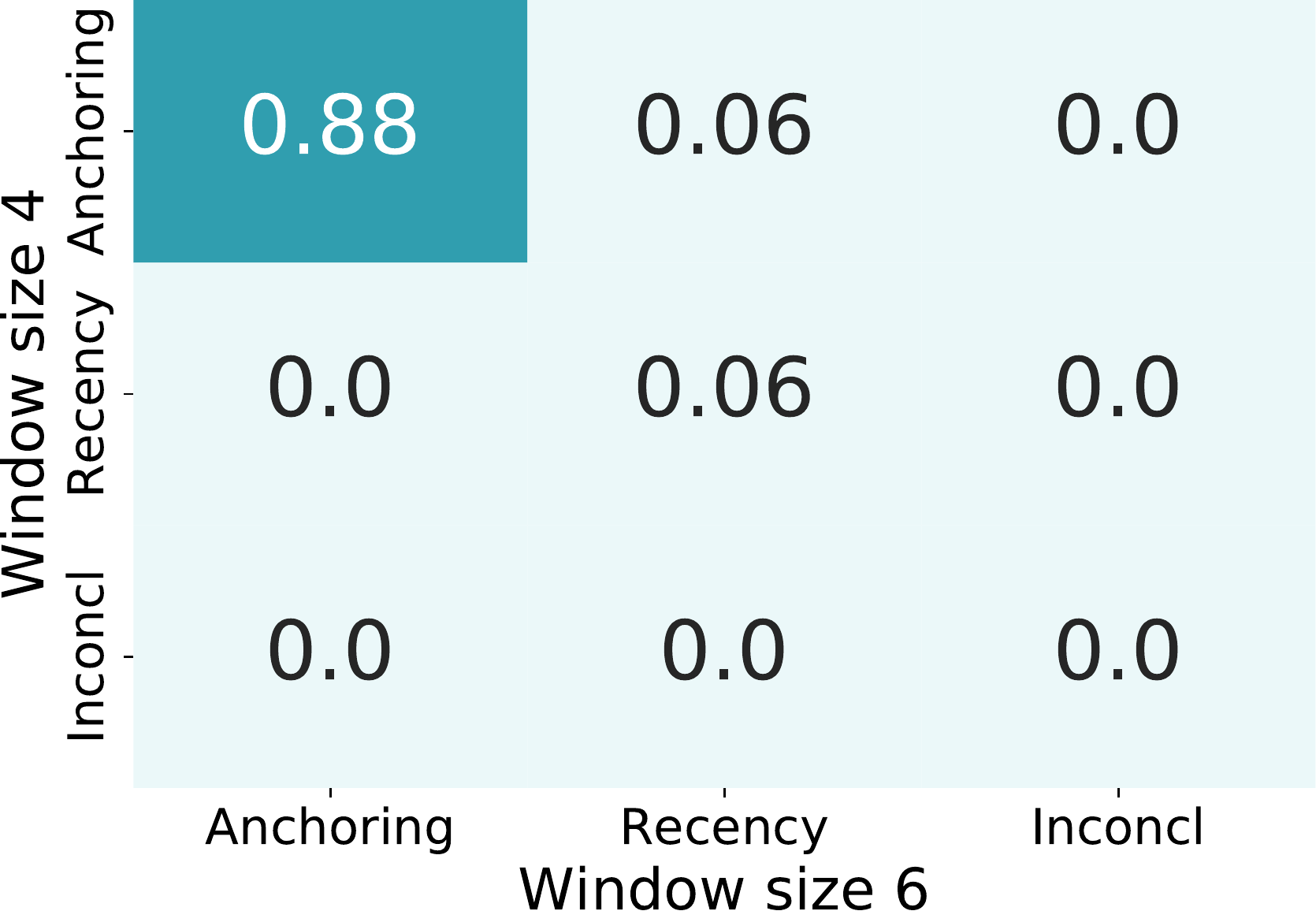}
     \end{subfigure}
     \begin{subfigure}
     \centering
     \includegraphics[width=0.27\textwidth]{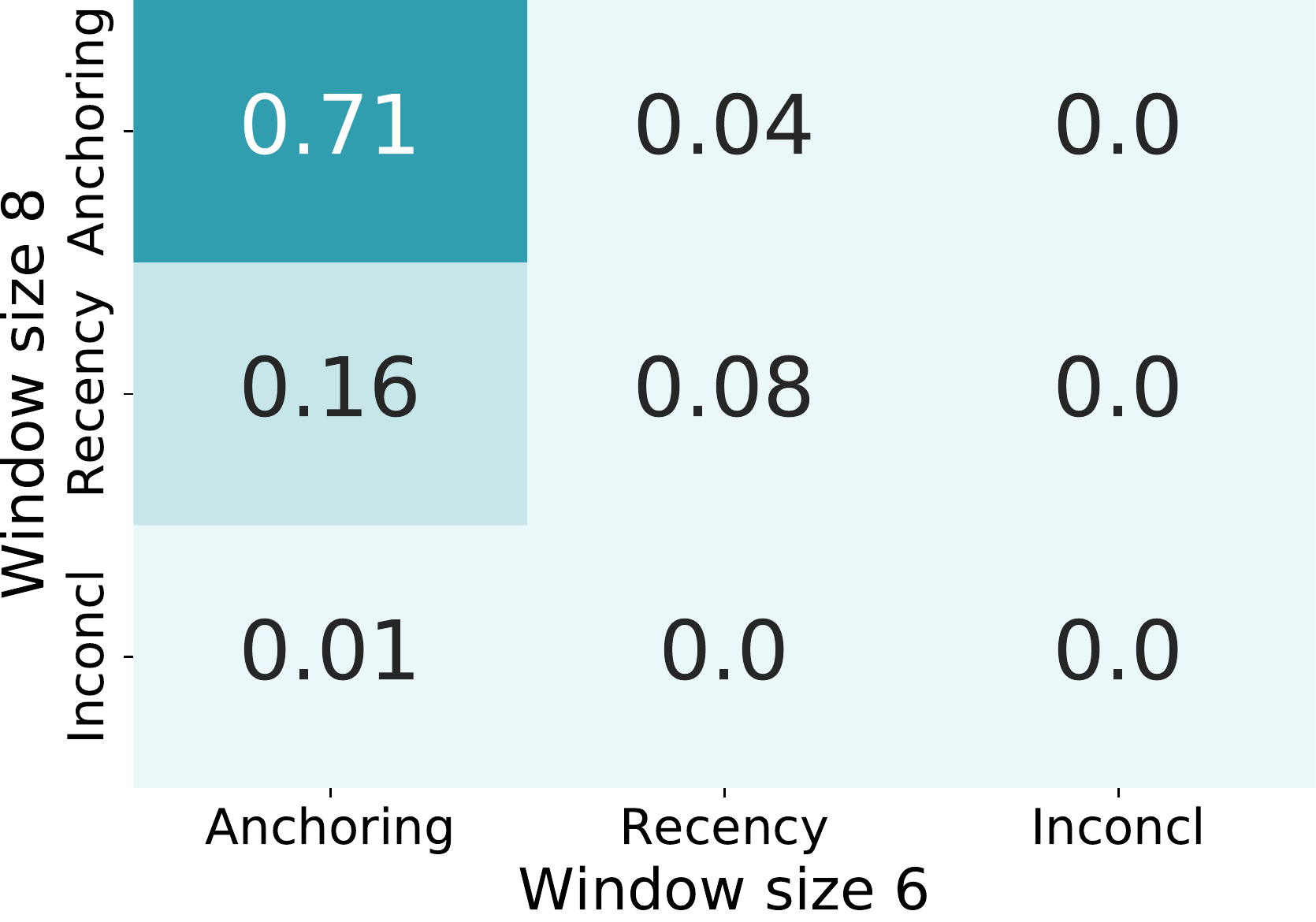}
     \end{subfigure}
     \begin{subfigure}
     \centering
     \includegraphics[width=0.27\textwidth]{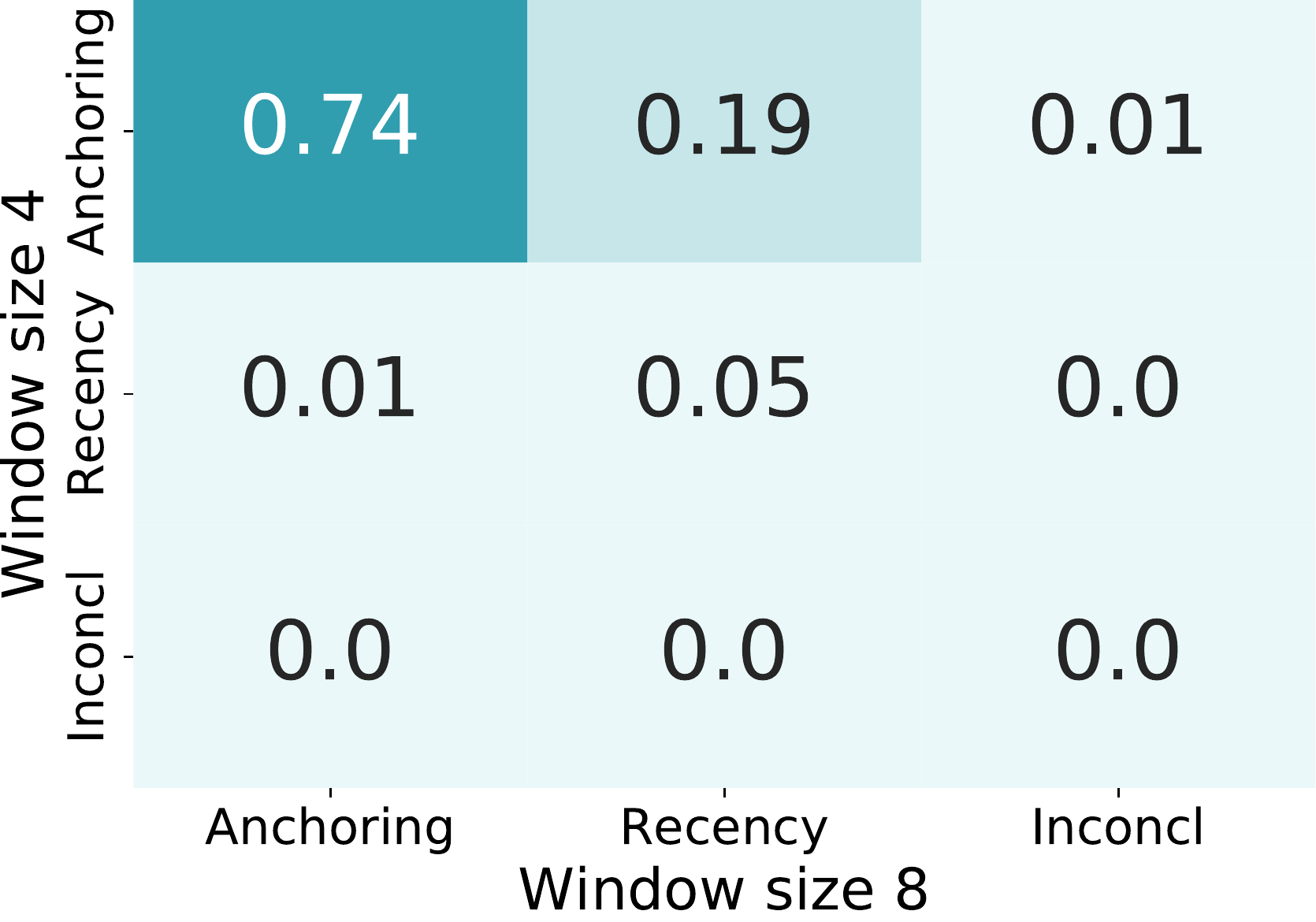}
     \end{subfigure}
     \caption{Detection-Sensitivity. Classification matrices for pairwise window sizes: (a) 4 \& 6, (b) 8 \& 6, (c) 4 \& 8. Pairwise-concordance scores 0.94, 0.79, 0.79, respectively, show sensitivity of detection to window sizes.}
     \label{compare}
\end{figure*}

\subsection{Detection-Reliability Experiment}\label{subsection_experiment_2}
We test whether the trend in deviations $\delta_{k,i}$ is consistent over a longer set of windows. 
The \textit{Extended-personalized} training, performed on data of each user, outputs attention weights for 6 visits, for each window of visits. 
The number of windows is different across users. For illustration, a randomly selected user's $\delta_{k,i}$ values are shown in \figurename ~\ref{attn_wt_3d}. The $\delta_{k,i}$ values (Z-axis) decrease over visits 1 to 6 (X-axis), and importantly, the trend in decreasing $\delta_{k,i}$ values over 6 visits is consistent across 17 windows (Y-axis), lending reliability to detection of anchoring bias for this user. 

This intuitive notion is put to a formal reliability test through a set of regression analyses. Since the goal is individual level detection, we run a separate regression for each user, as follows.
Deviation, $\delta_{k,i} = f(\beta_{k,1} x_{k,1,i}, \beta_{k,2} x_{k,2,i})$, 
where, $x_{k,1,i}$ and $x_{k,2,i}$ are visit-sequence number and window-sequence number, respectively, of user $k$ for visit $i$. The index $k$ for $\beta$ indicates user specific regression coefficient. The error follows N(0,$\sigma^2$). The quantity of interest, slope of $\delta_{k,i}$ over visit-sequence $x_{k,1,i}$ is given by $\beta_{k,1}$. Notably we account for the effect of window sequence $x_{k,2,i}$ in estimating $\beta_{k,1}$.
Each user's $\beta_{k,1}$ is tested statistically for $<$0 (anchoring), $>$0 (recency), or $=$0 (inconclusive). Table~\ref{tbl-p-value} shows that at $\alpha$-significance =0.05, across 244 users, the break-up is: $79\%$ indicate anchoring bias, $12\%$ recency bias and $9\%$ are inconclusive. Comparing the $9\%$ of inconclusive detection in our model  (Table~\ref{tbl-p-value}) with the $68\%$ to $72\%$ of inconclusiveness in the baseline (Table~\ref{tbl-freq-actions}), we conclude that the proposed HAN based approach using sequential action considerably outperforms the use of frequency distribution of actions in detecting these two biases.

\iffalse
\begin{figure}
     \centering
     \subfigure{\includegraphics[scale=0.23]{Images/SB_m_11_w46.pdf}}
     \subfigure{\includegraphics[scale=0.23]{Images/SB_m_11_w68.pdf}}
     \subfigure{\includegraphics[scale=0.23]{Images/SB_m_11_w48.pdf}}
     \caption{Detection-Sensitivity. Classification matrices for pairwise window sizes: (a) 4 \& 6, (b) 8 \& 6, (c) 4 \& 8. Pairwise-concordance scores 0.94, 0.79, 0.79, respectively, show sensitivity of detection to window sizes.}
     \label{compare}
\vskip -0.2in
\end{figure}
\fi

\subsection{Detection-Sensitivity Experiments}\label{subsection_experiment_3}
It is instructive to check whether users have the same bias detected by varying the number of visits within window, or window size. This yields sensitivity of the approach to our selection of window size.  
Three experiments are run using window sizes of 4, 6, 8. For like comparison across three window sizes and to meet the need for sufficient number of windows available per user for estimation, especially for sufficient number of windows with 8 visits each (which reduces the set of users), the data restricts to 97 users, and action set to 11 actions, (\textit{CS -/+ 5}); i.e., the context is 5 actions before and after \textit{CS}, in each visit. As before, each experiment performs common training to compute norm, followed by personalized training to compute individual's deviation $\delta_{k,i}$ from norm. Deviations $\delta_{k,i}$ are analyzed for individual bias detection. Classification is done at statistical $\alpha$-significance 0.05, to detect anchoring (slope$<$0) and recency (slope$>$0) biases. We use a pairwise concordant score, formally defined as follows. Assign label $k_q \in \{Anchoring, Inconclusive, Recency\}$ for user $k$ under experiment $q$, where $q$ refers to window size 4, 6, or 8.
Experiments $q$ and $q'$ yield pairwise concordance for user $k$ if the label for user $k$ is same under $q$ and $q'$. Thus,
\begin{equation}
    CP_{k} =
\begin{cases}
    1, & \textrm{if} \; k_q=k_{q'} \textrm{ for } \; q\neq q' \\
    0, & \textrm{otherwise}
\end{cases}
\end{equation}
The pairwise concordance score for all $K$ users is given by 
\begin{equation}
    CP = \frac{\sum_{k=1}^K CP_k}{K}
\end{equation}

Diagonal scores of matrices in \figurename ~\ref{compare} show concordance. Classification of biases has high concordance score, $0.94$, across window sizes 4$\times$6. Concordance scores for window-sizes 8$\times$6, and 4$\times$8, respectively, $0.79$, and $0.79$, are lower, although high on an absolute measure (1 is the max). The higher mis-classification is for window size 8. 

\iffalse
\subsection{Evaluation by Experts}
%To complement our interpretation of weights from HAN, 
As a qualitative validation step for the patterns of deviation that define these two biases (section~\ref{subsection_intuition_rationale}), an evaluation by experts, who are experienced analysts, is performed. No input from experts enter the model or analysis. With considerable effort, we got nine experts, who gave their interpretation of the patterns of deviation. To every expert three separate graphs were shown, each constructed from \figurename ~\ref{relative_attn}. Each graph showed three lines: graph 1 for flat pattern, graph 2 for decreasing pattern, and graph 3 for increasing pattern. For a deeper exposition of the task to these experts, we also provided them with more context by showing them several tables of sequences of actions from behavior logs, across many visits. With each graph shown separately, experts responded by selecting one option out of four possible options: (i) neither anchoring nor recency, (ii) anchoring, (iii) recency, (iv) cannot tell. Increasing weights were associated with recency bias by 8 out of 9 experts, confirming our interpretation for recency bias. Out of 9 experts, decreasing weights were associated with anchoring bias by 4, while remaining 5 selected among the responses (i) or (iv) or did not answer, providing partial support for our interpretation of anchoring bias. Experts corroborate the interpretation of pattern of deviations for bias detection. 
\fi

\section{Conclusion}
Cognitive biases occur in how the mind processes information, but manifest in actions and behaviors of humans, unbeknownst to themselves. Users are not aware of their biased actions while performing tasks that require objectivity. The impact of cognitive biases is wide ranging across fairness, accountability, transparency, ethics, law, medicine and discrimination, as discussed in the Introduction. 
Actions are observable, but how the mind processes information is unobserved. Thus, we set out to detect biases from actions. We confine to two biases - anchoring and recency. With behavior log data at our disposal we turn toward detection of these two biases from actions. The computer science literature is limited in offering ML methods for bias detection.  
We address three issues: (1) Introduce a principled framework to detect and distinguish between these biases from action level data, by using two principles \textit{PN1} and \textit{PN2} from cognitive psychology~\cite{hilbert2012toward}. (2) Offer individual level detection from behavior log to make detection valuable for each user. (3) Not rely upon annotated data. 

Modifying the training of a hierarchical attention network on behavior log we then interpret attention weights in a novel manner, to conform with proposed properties of deviation, for detecting and distinguishing between these two biases. From principle \textit{PN1}, we introduce a mathematical formulation of deviation in attention weights between common training and personalized training, to detect individual user specific bias. We offer two statistical approaches for testing. Our reliability experiment shows consistent detection across time, in line with principle \textit{PN2}. Additional experiments justify our approach by showing good paired-concordance in detection as window size and number of windows vary. We take a small step toward ML based individualized cognitive bias detection, a necessary step for bias mitigation. We examine only two out of many cognitive biases. We hope more AI and ML research examine cognitive biases from observable data to heighten awareness about these biases in humans performing objective tasks.


\bibliographystyle{ACM-Reference-Format}
\bibliography{references}


\begin{thebibliography}{38}

%%% the Web address.

\ifx \showCODEN    \undefined \def \showCODEN     #1{\unskip}     \fi
\ifx \showDOI      \undefined \def \showDOI       #1{#1}\fi
\ifx \showISBNx    \undefined \def \showISBNx     #1{\unskip}     \fi
\ifx \showISBNxiii \undefined \def \showISBNxiii  #1{\unskip}     \fi
\ifx \showISSN     \undefined \def \showISSN      #1{\unskip}     \fi
\ifx \showLCCN     \undefined \def \showLCCN      #1{\unskip}     \fi
\ifx \shownote     \undefined \def \shownote      #1{#1}          \fi
\ifx \showarticletitle \undefined \def \showarticletitle #1{#1}   \fi
\ifx \showURL      \undefined \def \showURL       {\relax}        \fi
% invisible to TeX
\providecommand\bibfield[2]{#2}
\providecommand\bibinfo[2]{#2}
\providecommand\natexlab[1]{#1}
\providecommand\showeprint[2][]{arXiv:#2}

\bibitem[\protect\citeauthoryear{Amini, Soleimany, Schwarting, Bhatia, and
  Rus}{Amini et~al\mbox{.}}{2019}]%
        {amini2019uncovering}
\bibfield{author}{\bibinfo{person}{Alexander Amini}, \bibinfo{person}{Ava~P
  Soleimany}, \bibinfo{person}{Wilko Schwarting}, \bibinfo{person}{Sangeeta~N
  Bhatia}, {and} \bibinfo{person}{Daniela Rus}.}
  \bibinfo{year}{2019}\natexlab{}.
\newblock \showarticletitle{Uncovering and mitigating algorithmic bias through
  learned latent structure}. In \bibinfo{booktitle}{\emph{Proceedings of the
  2019 AAAI/ACM Conference on AI, Ethics, and Society}}.
  \bibinfo{pages}{289--295}.
\newblock


\bibitem[\protect\citeauthoryear{Baeza-Yates}{Baeza-Yates}{2018}]%
        {baeza2018bias}
\bibfield{author}{\bibinfo{person}{Ricardo Baeza-Yates}.}
  \bibinfo{year}{2018}\natexlab{}.
\newblock \showarticletitle{Bias on the web}.
\newblock \bibinfo{journal}{\emph{Commun. ACM}} \bibinfo{volume}{61},
  \bibinfo{number}{6} (\bibinfo{year}{2018}), \bibinfo{pages}{54--61}.
\newblock


\bibitem[\protect\citeauthoryear{Bahdanau, Cho, and Bengio}{Bahdanau
  et~al\mbox{.}}{2016}]%
        {attention:1}
\bibfield{author}{\bibinfo{person}{Dzmitry Bahdanau},
  \bibinfo{person}{Kyunghyun Cho}, {and} \bibinfo{person}{Yoshua Bengio}.}
  \bibinfo{year}{2016}\natexlab{}.
\newblock \showarticletitle{Neural machine translation by jointly learning to
  align and translate}.
\newblock \bibinfo{journal}{\emph{arXiv:1409.0473 [cs.CL]}}
  (\bibinfo{year}{2016}).
\newblock


\bibitem[\protect\citeauthoryear{Barberis and Thaler}{Barberis and
  Thaler}{2003}]%
        {barberis2003survey}
\bibfield{author}{\bibinfo{person}{Nicholas Barberis} {and}
  \bibinfo{person}{Richard Thaler}.} \bibinfo{year}{2003}\natexlab{}.
\newblock \showarticletitle{A survey of behavioral finance}.
\newblock \bibinfo{journal}{\emph{Handbook of the Economics of Finance}}
  \bibinfo{volume}{1} (\bibinfo{year}{2003}), \bibinfo{pages}{1053--1128}.
\newblock


\bibitem[\protect\citeauthoryear{Bostrom and Ord}{Bostrom and Ord}{2006}]%
        {bostrom2006reversal}
\bibfield{author}{\bibinfo{person}{Nick Bostrom} {and} \bibinfo{person}{Toby
  Ord}.} \bibinfo{year}{2006}\natexlab{}.
\newblock \showarticletitle{The reversal test: eliminating status quo bias in
  applied ethics}.
\newblock \bibinfo{journal}{\emph{Ethics}} \bibinfo{volume}{116},
  \bibinfo{number}{4} (\bibinfo{year}{2006}), \bibinfo{pages}{656--679}.
\newblock


\bibitem[\protect\citeauthoryear{Croskerry}{Croskerry}{2013}]%
        {croskerry2013mindless}
\bibfield{author}{\bibinfo{person}{Pat Croskerry}.}
  \bibinfo{year}{2013}\natexlab{}.
\newblock \showarticletitle{From mindless to mindful practice—cognitive bias
  and clinical decision making}.
\newblock \bibinfo{journal}{\emph{N Engl J Med}} \bibinfo{volume}{368},
  \bibinfo{number}{26} (\bibinfo{year}{2013}), \bibinfo{pages}{2445--2448}.
\newblock


\bibitem[\protect\citeauthoryear{Edwards}{Edwards}{1954}]%
        {edwards1954theory}
\bibfield{author}{\bibinfo{person}{Ward Edwards}.}
  \bibinfo{year}{1954}\natexlab{}.
\newblock \showarticletitle{The theory of decision making.}
\newblock \bibinfo{journal}{\emph{Psychological bulletin}}
  \bibinfo{volume}{51}, \bibinfo{number}{4} (\bibinfo{year}{1954}),
  \bibinfo{pages}{380}.
\newblock


\bibitem[\protect\citeauthoryear{Eskridge~Jr and Ferejohn}{Eskridge~Jr and
  Ferejohn}{2001}]%
        {eskridge2001structuring}
\bibfield{author}{\bibinfo{person}{William~N Eskridge~Jr} {and}
  \bibinfo{person}{John Ferejohn}.} \bibinfo{year}{2001}\natexlab{}.
\newblock \showarticletitle{Structuring lawmaking to reduce cognitive bias: A
  critical view}.
\newblock \bibinfo{journal}{\emph{Cornell L. Rev.}}  \bibinfo{volume}{87}
  (\bibinfo{year}{2001}), \bibinfo{pages}{616}.
\newblock


\bibitem[\protect\citeauthoryear{Fortunato, Flammini, Menczer, and
  Vespignani}{Fortunato et~al\mbox{.}}{2006}]%
        {fortunato2006topical}
\bibfield{author}{\bibinfo{person}{Santo Fortunato},
  \bibinfo{person}{Alessandro Flammini}, \bibinfo{person}{Filippo Menczer},
  {and} \bibinfo{person}{Alessandro Vespignani}.}
  \bibinfo{year}{2006}\natexlab{}.
\newblock \showarticletitle{Topical interests and the mitigation of search
  engine bias}.
\newblock \bibinfo{journal}{\emph{Proceedings of the national academy of
  sciences}} \bibinfo{volume}{103}, \bibinfo{number}{34}
  (\bibinfo{year}{2006}), \bibinfo{pages}{12684--12689}.
\newblock


\bibitem[\protect\citeauthoryear{Gao and Shah}{Gao and Shah}{2021}]%
        {gao2021addressing}
\bibfield{author}{\bibinfo{person}{Ruoyuan Gao} {and} \bibinfo{person}{Chirag
  Shah}.} \bibinfo{year}{2021}\natexlab{}.
\newblock \showarticletitle{Addressing bias and fairness in search systems}. In
  \bibinfo{booktitle}{\emph{Proceedings of the 44th International ACM SIGIR
  Conference on Research and Development in Information Retrieval}}.
  \bibinfo{pages}{2643--2646}.
\newblock


\bibitem[\protect\citeauthoryear{Haselton, Nettle, and Murray}{Haselton
  et~al\mbox{.}}{2015}]%
        {haselton2015evolution}
\bibfield{author}{\bibinfo{person}{Martie~G Haselton}, \bibinfo{person}{Daniel
  Nettle}, {and} \bibinfo{person}{Damian~R Murray}.}
  \bibinfo{year}{2015}\natexlab{}.
\newblock \showarticletitle{The evolution of cognitive bias}.
\newblock \bibinfo{journal}{\emph{The handbook of evolutionary psychology}}
  (\bibinfo{year}{2015}), \bibinfo{pages}{1--20}.
\newblock


\bibitem[\protect\citeauthoryear{Hilbert}{Hilbert}{2012}]%
        {hilbert2012toward}
\bibfield{author}{\bibinfo{person}{Martin Hilbert}.}
  \bibinfo{year}{2012}\natexlab{}.
\newblock \showarticletitle{Toward a synthesis of cognitive biases: how noisy
  information processing can bias human decision making.}
\newblock \bibinfo{journal}{\emph{Psychological bulletin}}
  \bibinfo{volume}{138}, \bibinfo{number}{2} (\bibinfo{year}{2012}),
  \bibinfo{pages}{211}.
\newblock


\bibitem[\protect\citeauthoryear{Ieong, Mishra, Sadikov, and Zhang}{Ieong
  et~al\mbox{.}}{2012}]%
        {ieong2012domain}
\bibfield{author}{\bibinfo{person}{Samuel Ieong}, \bibinfo{person}{Nina
  Mishra}, \bibinfo{person}{Eldar Sadikov}, {and} \bibinfo{person}{Li Zhang}.}
  \bibinfo{year}{2012}\natexlab{}.
\newblock \showarticletitle{Domain bias in web search}. In
  \bibinfo{booktitle}{\emph{Proceedings of the fifth ACM international
  conference on Web search and data mining}}. \bibinfo{pages}{413--422}.
\newblock


\bibitem[\protect\citeauthoryear{Joachims, Granka, Pan, Hembrooke, Radlinski,
  and Gay}{Joachims et~al\mbox{.}}{2007}]%
        {joachims2007evaluating}
\bibfield{author}{\bibinfo{person}{Thorsten Joachims}, \bibinfo{person}{Laura
  Granka}, \bibinfo{person}{Bing Pan}, \bibinfo{person}{Helene Hembrooke},
  \bibinfo{person}{Filip Radlinski}, {and} \bibinfo{person}{Geri Gay}.}
  \bibinfo{year}{2007}\natexlab{}.
\newblock \showarticletitle{Evaluating the accuracy of implicit feedback from
  clicks and query reformulations in web search}.
\newblock \bibinfo{journal}{\emph{ACM Transactions on Information Systems
  (TOIS)}} \bibinfo{volume}{25}, \bibinfo{number}{2} (\bibinfo{year}{2007}),
  \bibinfo{pages}{7--es}.
\newblock


\bibitem[\protect\citeauthoryear{Jones and Skarlicki}{Jones and
  Skarlicki}{2013}]%
        {jones2013perceptions}
\bibfield{author}{\bibinfo{person}{David~A Jones} {and}
  \bibinfo{person}{Daniel~P Skarlicki}.} \bibinfo{year}{2013}\natexlab{}.
\newblock \showarticletitle{How perceptions of fairness can change: A dynamic
  model of organizational justice}.
\newblock \bibinfo{journal}{\emph{Organizational psychology review}}
  \bibinfo{volume}{3}, \bibinfo{number}{2} (\bibinfo{year}{2013}),
  \bibinfo{pages}{138--160}.
\newblock


\bibitem[\protect\citeauthoryear{Jordan and Mitchell}{Jordan and
  Mitchell}{2015}]%
        {jordan2015machine}
\bibfield{author}{\bibinfo{person}{Michael~I Jordan} {and}
  \bibinfo{person}{Tom~M Mitchell}.} \bibinfo{year}{2015}\natexlab{}.
\newblock \showarticletitle{Machine learning: Trends, perspectives, and
  prospects}.
\newblock \bibinfo{journal}{\emph{Science}} \bibinfo{volume}{349},
  \bibinfo{number}{6245} (\bibinfo{year}{2015}), \bibinfo{pages}{255--260}.
\newblock


\bibitem[\protect\citeauthoryear{Kliegr, Bahn{\'\i}k, and F{\"u}rnkranz}{Kliegr
  et~al\mbox{.}}{2018}]%
        {kliegr2018review}
\bibfield{author}{\bibinfo{person}{Tom{\'a}{\v{s}} Kliegr},
  \bibinfo{person}{{\v{S}}t{\v{e}}p{\'a}n Bahn{\'\i}k}, {and}
  \bibinfo{person}{Johannes F{\"u}rnkranz}.} \bibinfo{year}{2018}\natexlab{}.
\newblock \showarticletitle{A review of possible effects of cognitive biases on
  interpretation of rule-based machine learning models}.
\newblock \bibinfo{journal}{\emph{arXiv preprint arXiv:1804.02969}}
  (\bibinfo{year}{2018}).
\newblock


\bibitem[\protect\citeauthoryear{Krieger}{Krieger}{1995}]%
        {krieger1995content}
\bibfield{author}{\bibinfo{person}{Linda~Hamilton Krieger}.}
  \bibinfo{year}{1995}\natexlab{}.
\newblock \showarticletitle{The content of our categories: A cognitive bias
  approach to discrimination and equal employment opportunity}.
\newblock \bibinfo{journal}{\emph{Stanford Law Review}} (\bibinfo{year}{1995}),
  \bibinfo{pages}{1161--1248}.
\newblock


\bibitem[\protect\citeauthoryear{Lerner and Tetlock}{Lerner and
  Tetlock}{1999}]%
        {lerner1999accounting}
\bibfield{author}{\bibinfo{person}{Jennifer~S Lerner} {and}
  \bibinfo{person}{Philip~E Tetlock}.} \bibinfo{year}{1999}\natexlab{}.
\newblock \showarticletitle{Accounting for the effects of accountability.}
\newblock \bibinfo{journal}{\emph{Psychological bulletin}}
  \bibinfo{volume}{125}, \bibinfo{number}{2} (\bibinfo{year}{1999}),
  \bibinfo{pages}{255}.
\newblock


\bibitem[\protect\citeauthoryear{Mehrabi, Morstatter, Saxena, Lerman, and
  Galstyan}{Mehrabi et~al\mbox{.}}{2019}]%
        {mehrabi2019survey}
\bibfield{author}{\bibinfo{person}{Ninareh Mehrabi}, \bibinfo{person}{Fred
  Morstatter}, \bibinfo{person}{Nripsuta Saxena}, \bibinfo{person}{Kristina
  Lerman}, {and} \bibinfo{person}{Aram Galstyan}.}
  \bibinfo{year}{2019}\natexlab{}.
\newblock \showarticletitle{A survey on bias and fairness in machine learning}.
\newblock \bibinfo{journal}{\emph{arXiv preprint arXiv:1908.09635}}
  (\bibinfo{year}{2019}).
\newblock


\bibitem[\protect\citeauthoryear{Pompian}{Pompian}{2011}]%
        {pompian2011behavioral}
\bibfield{author}{\bibinfo{person}{Michael~M Pompian}.}
  \bibinfo{year}{2011}\natexlab{}.
\newblock \bibinfo{booktitle}{\emph{Behavioral finance and wealth management:
  how to build investment strategies that account for investor biases}}.
  Vol.~\bibinfo{volume}{667}.
\newblock \bibinfo{publisher}{John Wiley \& Sons}.
\newblock


\bibitem[\protect\citeauthoryear{Pothirattanachaikul, Yamamoto, Yamamoto, and
  Yoshikawa}{Pothirattanachaikul et~al\mbox{.}}{2020}]%
        {pothirattanachaikul2020analyzing}
\bibfield{author}{\bibinfo{person}{Suppanut Pothirattanachaikul},
  \bibinfo{person}{Takehiro Yamamoto}, \bibinfo{person}{Yusuke Yamamoto}, {and}
  \bibinfo{person}{Masatoshi Yoshikawa}.} \bibinfo{year}{2020}\natexlab{}.
\newblock \showarticletitle{Analyzing the Effects of" People also ask" on
  Search Behaviors and Beliefs}. In \bibinfo{booktitle}{\emph{Proceedings of
  the 31st ACM Conference on Hypertext and Social Media}}.
  \bibinfo{pages}{101--110}.
\newblock


\bibitem[\protect\citeauthoryear{Rabin}{Rabin}{1998}]%
        {rabin1998psychology}
\bibfield{author}{\bibinfo{person}{Matthew Rabin}.}
  \bibinfo{year}{1998}\natexlab{}.
\newblock \showarticletitle{Psychology and economics}.
\newblock \bibinfo{journal}{\emph{Journal of economic literature}}
  \bibinfo{volume}{36}, \bibinfo{number}{1} (\bibinfo{year}{1998}),
  \bibinfo{pages}{11--46}.
\newblock


\bibitem[\protect\citeauthoryear{Rabin and Schrag}{Rabin and Schrag}{1999}]%
        {rabin1999first}
\bibfield{author}{\bibinfo{person}{Matthew Rabin} {and} \bibinfo{person}{Joel~L
  Schrag}.} \bibinfo{year}{1999}\natexlab{}.
\newblock \showarticletitle{First impressions matter: A model of confirmatory
  bias}.
\newblock \bibinfo{journal}{\emph{The quarterly journal of economics}}
  \bibinfo{volume}{114}, \bibinfo{number}{1} (\bibinfo{year}{1999}),
  \bibinfo{pages}{37--82}.
\newblock


\bibitem[\protect\citeauthoryear{Ramdas, Trillos, and Cuturi}{Ramdas
  et~al\mbox{.}}{2017}]%
        {Ramdas_2017}
\bibfield{author}{\bibinfo{person}{Aaditya Ramdas}, \bibinfo{person}{Nicolás
  Trillos}, {and} \bibinfo{person}{Marco Cuturi}.}
  \bibinfo{year}{2017}\natexlab{}.
\newblock \showarticletitle{On Wasserstein Two-Sample Testing and Related
  Families of Nonparametric Tests}.
\newblock \bibinfo{journal}{\emph{Entropy}} \bibinfo{volume}{19},
  \bibinfo{number}{2} (\bibinfo{date}{Jan} \bibinfo{year}{2017}),
  \bibinfo{pages}{47}.
\newblock
\showISSN{1099-4300}
\urldef\tempurl%
\url{https://doi.org/10.3390/e19020047}
\showDOI{\tempurl}


\bibitem[\protect\citeauthoryear{Schaerer, Kern, Berger, Medvec, and
  Swaab}{Schaerer et~al\mbox{.}}{2018}]%
        {schaerer2018illusion}
\bibfield{author}{\bibinfo{person}{Michael Schaerer}, \bibinfo{person}{Mary
  Kern}, \bibinfo{person}{Gail Berger}, \bibinfo{person}{Victoria Medvec},
  {and} \bibinfo{person}{Roderick~I Swaab}.} \bibinfo{year}{2018}\natexlab{}.
\newblock \showarticletitle{The illusion of transparency in performance
  appraisals: When and why accuracy motivation explains unintentional feedback
  inflation}.
\newblock \bibinfo{journal}{\emph{Organizational Behavior and Human Decision
  Processes}}  \bibinfo{volume}{144} (\bibinfo{year}{2018}),
  \bibinfo{pages}{171--186}.
\newblock


\bibitem[\protect\citeauthoryear{Simon}{Simon}{1955}]%
        {simon1955behavioral}
\bibfield{author}{\bibinfo{person}{Herbert~A Simon}.}
  \bibinfo{year}{1955}\natexlab{}.
\newblock \showarticletitle{A behavioral model of rational choice}.
\newblock \bibinfo{journal}{\emph{The quarterly journal of economics}}
  \bibinfo{volume}{69}, \bibinfo{number}{1} (\bibinfo{year}{1955}),
  \bibinfo{pages}{99--118}.
\newblock


\bibitem[\protect\citeauthoryear{Simon}{Simon}{1956}]%
        {simon1956rational}
\bibfield{author}{\bibinfo{person}{Herbert~A Simon}.}
  \bibinfo{year}{1956}\natexlab{}.
\newblock \showarticletitle{Rational choice and the structure of the
  environment.}
\newblock \bibinfo{journal}{\emph{Psychological review}} \bibinfo{volume}{63},
  \bibinfo{number}{2} (\bibinfo{year}{1956}), \bibinfo{pages}{129}.
\newblock


\bibitem[\protect\citeauthoryear{Song, Wang, and He}{Song
  et~al\mbox{.}}{2014}]%
        {10.1145/2556195.2556234}
\bibfield{author}{\bibinfo{person}{Yang Song}, \bibinfo{person}{Hongning Wang},
  {and} \bibinfo{person}{Xiaodong He}.} \bibinfo{year}{2014}\natexlab{}.
\newblock \showarticletitle{Adapting Deep RankNet for Personalized Search}. In
  \bibinfo{booktitle}{\emph{Proceedings of the 7th ACM International Conference
  on Web Search and Data Mining}} (New York, New York, USA)
  \emph{(\bibinfo{series}{WSDM ’14})}. \bibinfo{publisher}{Association for
  Computing Machinery}, \bibinfo{address}{New York, NY, USA},
  \bibinfo{pages}{83–92}.
\newblock
\showISBNx{9781450323512}
\urldef\tempurl%
\url{https://doi.org/10.1145/2556195.2556234}
\showDOI{\tempurl}


\bibitem[\protect\citeauthoryear{Stroessner and Heuer}{Stroessner and
  Heuer}{1996}]%
        {stroessner1996cognitive}
\bibfield{author}{\bibinfo{person}{Steven~J Stroessner} {and}
  \bibinfo{person}{Larry~B Heuer}.} \bibinfo{year}{1996}\natexlab{}.
\newblock \showarticletitle{Cognitive bias in procedural justice: Formation and
  implications of illusory correlations in perceived intergroup fairness.}
\newblock \bibinfo{journal}{\emph{Journal of Personality and Social
  Psychology}} \bibinfo{volume}{71}, \bibinfo{number}{4}
  (\bibinfo{year}{1996}), \bibinfo{pages}{717}.
\newblock


\bibitem[\protect\citeauthoryear{Tetlock and Mitchell}{Tetlock and
  Mitchell}{2009}]%
        {tetlock2009implicit}
\bibfield{author}{\bibinfo{person}{Philip~E Tetlock} {and}
  \bibinfo{person}{Gregory Mitchell}.} \bibinfo{year}{2009}\natexlab{}.
\newblock \showarticletitle{Implicit bias and accountability systems: What must
  organizations do to prevent discrimination?}
\newblock \bibinfo{journal}{\emph{Research in organizational behavior}}
  \bibinfo{volume}{29} (\bibinfo{year}{2009}), \bibinfo{pages}{3--38}.
\newblock


\bibitem[\protect\citeauthoryear{Tversky and Kahneman}{Tversky and
  Kahneman}{1974}]%
        {tversky1974judgment}
\bibfield{author}{\bibinfo{person}{Amos Tversky} {and} \bibinfo{person}{Daniel
  Kahneman}.} \bibinfo{year}{1974}\natexlab{}.
\newblock \showarticletitle{Judgment under uncertainty: Heuristics and biases}.
\newblock \bibinfo{journal}{\emph{science}} \bibinfo{volume}{185},
  \bibinfo{number}{4157} (\bibinfo{year}{1974}), \bibinfo{pages}{1124--1131}.
\newblock


\bibitem[\protect\citeauthoryear{Wang, Wang, Tang, O'Hare, Chang, and Li}{Wang
  et~al\mbox{.}}{2016}]%
        {DBLP:journals/corr/WangWTOCL16}
\bibfield{author}{\bibinfo{person}{Yilin Wang}, \bibinfo{person}{Suhang Wang},
  \bibinfo{person}{Jiliang Tang}, \bibinfo{person}{Neil O'Hare},
  \bibinfo{person}{Yi Chang}, {and} \bibinfo{person}{Baoxin Li}.}
  \bibinfo{year}{2016}\natexlab{}.
\newblock \showarticletitle{Hierarchical Attention Network for Action
  Recognition in Videos}.
\newblock \bibinfo{journal}{\emph{CoRR}}  \bibinfo{volume}{abs/1607.06416}
  (\bibinfo{year}{2016}).
\newblock
\showeprint[arxiv]{1607.06416}
\urldef\tempurl%
\url{http://arxiv.org/abs/1607.06416}
\showURL{%
\tempurl}


\bibitem[\protect\citeauthoryear{White}{White}{2013}]%
        {white2013beliefs}
\bibfield{author}{\bibinfo{person}{Ryen White}.}
  \bibinfo{year}{2013}\natexlab{}.
\newblock \showarticletitle{Beliefs and biases in web search}. In
  \bibinfo{booktitle}{\emph{Proceedings of the 36th international ACM SIGIR
  conference on Research and development in information retrieval}}.
  \bibinfo{pages}{3--12}.
\newblock


\bibitem[\protect\citeauthoryear{White}{White}{2014}]%
        {white2014belief}
\bibfield{author}{\bibinfo{person}{Ryen~W White}.}
  \bibinfo{year}{2014}\natexlab{}.
\newblock \showarticletitle{Belief dynamics in Web search}.
\newblock \bibinfo{journal}{\emph{Journal of the Association for Information
  Science and Technology}} \bibinfo{volume}{65}, \bibinfo{number}{11}
  (\bibinfo{year}{2014}), \bibinfo{pages}{2165--2178}.
\newblock


\bibitem[\protect\citeauthoryear{White and Hassan}{White and Hassan}{2014}]%
        {white2014content}
\bibfield{author}{\bibinfo{person}{Ryen~W White} {and} \bibinfo{person}{Ahmed
  Hassan}.} \bibinfo{year}{2014}\natexlab{}.
\newblock \showarticletitle{Content bias in online health search}.
\newblock \bibinfo{journal}{\emph{ACM Transactions on the Web (TWEB)}}
  \bibinfo{volume}{8}, \bibinfo{number}{4} (\bibinfo{year}{2014}),
  \bibinfo{pages}{1--33}.
\newblock


\bibitem[\protect\citeauthoryear{White and Horvitz}{White and Horvitz}{2015}]%
        {white2015belief}
\bibfield{author}{\bibinfo{person}{Ryen~W White} {and} \bibinfo{person}{Eric
  Horvitz}.} \bibinfo{year}{2015}\natexlab{}.
\newblock \showarticletitle{Belief dynamics and biases in web search}.
\newblock \bibinfo{journal}{\emph{ACM Transactions on Information Systems
  (TOIS)}} \bibinfo{volume}{33}, \bibinfo{number}{4} (\bibinfo{year}{2015}),
  \bibinfo{pages}{1--46}.
\newblock


\bibitem[\protect\citeauthoryear{Yang, Yang, Dyer, He, Smola, and Hovy}{Yang
  et~al\mbox{.}}{2016}]%
        {hierarchical_attn:1}
\bibfield{author}{\bibinfo{person}{Zichao Yang}, \bibinfo{person}{Diyi Yang},
  \bibinfo{person}{Chris Dyer}, \bibinfo{person}{Xiaodong He},
  \bibinfo{person}{Alex Smola}, {and} \bibinfo{person}{Eduard Hovy}.}
  \bibinfo{year}{2016}\natexlab{}.
\newblock \showarticletitle{Hierarchical attention networks for document
  classification}.
\newblock \bibinfo{journal}{\emph{Proceedings of the 2016 conference of the
  North American chapter of the association for computational linguistics:
  human language technologies}} (\bibinfo{year}{2016}).
\newblock


\end{thebibliography}

\end{document}